\documentclass[manuscript,screen]{acmart}
\AtBeginDocument{%
  }

\setcopyright{acmlicensed}
\copyrightyear{2018}
\acmYear{2018}
\acmDOI{XXXXXXX.XXXXXXX}

\acmConference[Conference acronym 'XX]{Make sure to enter the correct
  conference title from your rights confirmation emai}{June 03--05,
  2018}{Woodstock, NY}
\acmISBN{978-1-4503-XXXX-X/18/06}

\usepackage{graphicx}
\usepackage{tikz}
\usepackage{forest}
\usetikzlibrary{trees,positioning,shapes,shadows,arrows.meta}
\usepackage{subcaption}
\usepackage{caption} 
\usepackage{hyperref}
\usepackage{xcolor}
\usepackage{pifont}
\usepackage{tabularx}
\usepackage{multirow}

\definecolor{arch_design}{HTML}{0072B2}
\definecolor{pre_training}{HTML}{E69F00}
\definecolor{fine_tuning}{HTML}{A2C814}
\definecolor{inference}{HTML}{F99AAE}
\definecolor{system_design}{HTML}{9DA3A6}
\definecolor{resource_label}{HTML}{C3B090}
\definecolor{light_blue}{HTML}{9AD0EC}
\definecolor{light_orange}{HTML}{F8C471}
\definecolor{light_green}{HTML}{C6ECAE}
\definecolor{light_pink}{HTML}{FAD9DC}
\definecolor{light_gray}{HTML}{D3D3D3}

\begin{document}

\title{A Survey on Inference Optimization Techniques for Mixture of Experts Models}

\author{Jiacheng Liu}
\authornote{Both authors contributed equally to this research.}
\email{jcliu@cse.cuhk.edu.hk}
\affiliation{%
  \institution{Chinese University of Hong Kong}
  \city{Hong Kong}
  \country{China}
}

\author{Peng Tang}
\authornotemark[1]
\affiliation{%
  \institution{Shanghai Jiao Tong Universtiy}
  \city{Shanghai}
  \country{China}}

\author{Wenfeng Wang}
\affiliation{%
  \institution{Shanghai Jiao Tong Universtiy}
  \city{Shanghai}
  \country{China}}

\author{Yuhang Ren}
\affiliation{%
  \institution{Shanghai Jiao Tong Universtiy}
  \city{Shanghai}
  \country{China}}

\author{Xiaofeng Hou}
\authornote{Corresponding authors.}
\affiliation{%
  \institution{Shanghai Jiao Tong Universtiy}
  \city{Shanghai}
  \country{China}}

\author{Pheng-Ann Heng}
\affiliation{%
  \institution{Chinese University of Hong Kong}
  \city{Hong Kong}
  \country{China}
}

\author{Minyi Guo}
\affiliation{%
  \institution{Shanghai Jiao Tong Universtiy}
  \city{Shanghai}
  \country{China}
}

\author{Chao Li}
\authornotemark[2]
\affiliation{%
  \institution{Shanghai Jiao Tong Universtiy}
  \city{Shanghai}
  \country{China}}

\renewcommand{\shortauthors}{Liu et al.}

\begin{abstract}
The emergence of large-scale Mixture of Experts (MoE) models represents a significant advancement in artificial intelligence, offering enhanced model capacity and computational efficiency through conditional computation. However, deploying and running inference on these models presents significant challenges in computational resources, latency, and energy efficiency. This comprehensive survey analyzes optimization techniques for MoE models across the entire system stack.
  We first establish a taxonomical framework that categorizes optimization approaches into model-level, system-level, and hardware-level optimizations. At the model level, we examine architectural innovations including efficient expert design, attention mechanisms, various compression techniques such as pruning, quantization, and knowledge distillation, as well as algorithm improvement including dynamic routing strategies and expert merging methods. At the system level, we investigate distributed computing approaches, load balancing mechanisms, and efficient scheduling algorithms that enable scalable deployment. Furthermore, we delve into hardware-specific optimizations and co-design strategies that maximize throughput and energy efficiency. 
This survey provides both a structured overview of existing solutions and identifies key challenges and promising research directions in MoE inference optimization. 
   To facilitate ongoing updates and the sharing of cutting-edge advances in MoE
inference optimization research, we have established a repository accessible at  \url{https://github.com/MoE-Inf/awesome-moe-inference/}.

\end{abstract}



\received{20 February 2007}
\received[revised]{12 March 2009}
\received[accepted]{5 June 2009}

\maketitle

\section{Introduction}
Large language models (LLMs) have revolutionized artificial intelligence, demonstrating unprecedented capabilities across various domains including natural language processing~\cite{vaswani2017attention, ouyang2022training, chowdhery2023palm}, computer vision~\cite{dosovitskiy2020image, pmlr-v202-dehghani23a, zhang2021vit}, and multimodal tasks~\cite{radford2021learning, li2021align, wang2022ofa}. Models like GPT-4~\cite{achiam2023gpt}, Claude~\cite{Claude}, and Gemini~\cite{team2024gemini} have achieved remarkable performance in tasks ranging from natural language understanding to complex reasoning and code generation. The impressive capabilities of these models are largely attributed to their massive scale, both in terms of model parameters and computational resources invested in training. This scaling trend is supported by empirical evidence showing consistent improvements in model performance with increased size, as demonstrated by various scaling laws in language modeling and other domains~\cite{alabdulmohsin2022revisiting, kaplan2020scaling, cherti2023reproducible}. However, this trajectory presents significant challenges in terms of computational efficiency and resource utilization, particularly during inference, where real-world deployment constraints become critical~\cite{zhou2024survey, xu2024survey, bai2024beyond, yuan2024llm}.

Mixture of Experts (MoE) has emerged as a promising architectural solution to address scaling challenges in machine learning~\cite{shazeer2017outrageously}. Originally introduced by Jacobs et al.\cite{articleJacobs} in the early 1990s as a method for learning subtasks in neural networks. Numerous MoE-based models~\cite{MetaPi, NIPS20009fdb62f9, Eigen2013LearningFR} have been developed over the years. In the era of large language models, MoE has again experienced a renaissance~\cite{jiang2024mixtral, abdin2024phi, dai2024deepseekmoe, sun2024hunyuan}.  The core principle of MoE is to distribute the model's capacity across multiple specialized sub-networks, or experts, with a learned routing mechanism that selectively activates only the relevant experts for each input. This approach allows models to maintain a large parameter count while keeping computational costs manageable through sparse activation. Recent implementations, such as Mixtral 8x7B~\cite{jiang2024mixtral}, DeepSeek-V3~\cite{liu2024deepseek} and DBRX~\cite{DBRX}, have demonstrated the effectiveness of this strategy in scaling language models to trillions of parameters while maintaining reasonable computational requirements.

\begin{table}[t]
\centering
\begin{tabular}{cccccccccc}
\hline

\textbf{Reference} & \textbf{Para.} & \textbf{Experts} & \textbf{\#L} & \textbf{\#H} & \textbf{$d_{model}$} & \textbf{$d_{ffn}$} & \textbf{$d_{expert}$} & \textbf{Affiliation} & \textbf{Time} \\
\hline

\href{https://huggingface.co/facebook/nllb-moe-54b}{NLLB~\cite{costa2022no}} & 54B & 2/64/0 & 24 & 16 & 1024 & 8192 & 8192 & FaceBook & 2022.07 \\
\hline

\href{https://huggingface.co/Qwen/Qwen2-57B-A14B}{Qwen2-57B-A14B~\cite{yang2024qwen2}} & 57.4B & 8/64/0 & 28 & 28 & 3584 & 18944 & 2560 & Alibaba & 2023.05 \\
\hline

\href{https://huggingface.co/mistralai/Mixtral-8x7B-v0.1}{Mixtral-8x7B~\cite{jiang2024mixtral}} & 46.7B & 2/8/0 & 32 & 32 & 4096 & 14336 & 14336 & Mistral AI & 2023.12 \\
\hline

\href{https://huggingface.co/OrionZheng/openmoe-base}{OpenMoE~\cite{xue2024openmoe}} & 34B & 2/16/0 & 12 & 12 & 768 & 2048 & 2048 & NUS et al. & 2023.12 \\
\hline

\href{https://huggingface.co/deepseek-ai/deepseek-moe-16b-base}{DeepSeekMoE~\cite{dai2024deepseekmoe}} & 16.4B & 6/64/2 & 28 & 16 & 2048 & 10944 & 1408 & DeepSeek-AI & 2024.01 \\
\hline

\href{https://huggingface.co/Qwen/Qwen1.5-MoE-A2.7B}{Qwen1.5-MoE~\cite{qwen_moe}} & 14.3B & 4/60/0 & 24 & 16 & 2048 & 5632 & 1408 & Alibaba & 2024.02 \\
\hline

\href{https://huggingface.co/jetmoe/jetmoe-8b}{JetMoE~\cite{shen2024jetmoe}} & 8.52B & 2/8/0 & 24 & 32 & 2048 & 5632 & 5632 & MIT et al. & 2024.03 \\
\hline

\href{https://huggingface.co/ai21labs/Jamba-v0.1}{Jamba~\cite{lieber2024jamba}} & 51.6B & 2/16/0 & 32 & 32 & 4096 & 14336 & 14336 & ai21labs & 2024.03 \\
\hline

\href{https://huggingface.co/databricks/dbrx-base}{DBRX~\cite{DBRX}} & 132B & 4/16/0 & 40 & 48 & 6144 & 10752 & 10752 & Databricks & 2024.03 \\
\hline

\href{https://huggingface.co/xai-org/grok-1}{Grok-1~\cite{Grok-1}} & 314B & 2/8/0 & 64 & 48 & 6144 & UNK & UNK & xAI & 2024.03 \\
\hline

\href{https://huggingface.co/Snowflake/snowflake-arctic-base}{Arctic~\cite{Arctic}} & 482B & 2/128/0 & 35 & 56 & 7168 & 4864 & 4864 & Snowflake & 2024.04 \\
\hline

\href{https://huggingface.co/mistralai/Mixtral-8x22B-v0.1}{Mixtral-8x22B~\cite{jiang2024mixtral}} & 141B & 2/8/0 & 56 & 48 & 6144 & 16384 & 16384 & Mistral AI & 2024.04 \\
\hline

\href{https://huggingface.co/deepseek-ai/DeepSeek-V2}{DeepSeek-V2~\cite{deepseekai2024deepseekv2strongeconomicalefficient}} & 236B & 6/160/2 & 60 & 128 & 5120 & 12288 & 1536 & DeepSeek-AI & 2024.04 \\
\hline

\href{https://huggingface.co/Skywork/Skywork-MoE-Base}{Skywork-MoE~\cite{wei2024skywork}} & 13B & 2/16/0 & 52 & 36 & 4608 & 12288 & 12288 & Kunlun Tech & 2024.05 \\
\hline

\href{https://huggingface.co/IEITYuan/Yuan2-M32-hf}{Yuan2~\cite{wu2024yuan}} & 40B & 2/32/0 & 24 & 16 & 2048 & 8192 & 8192 & IEIT-Yuan & 2024.05 \\
\hline

\href{https://github.com/pjlab-sys4nlp/llama-moe}{LLaMA-MoE~\cite{zhu2024llama}} & 6.7B & 2/8/0 & 32 & 32 & 4096 & 11008 & 11008 & Zhu et al. & 2024.06 \\
\hline

\href{https://huggingface.co/allenai/OLMoE-1B-7B-0924}{OLMoE~\cite{muennighoff2024olmoe}} & 6.92B & 8/64/0 & 16 & 16 & 2048 & 1024 & 1024 & AllenAI & 2024.07 \\
\hline

\href{https://huggingface.co/microsoft/Phi-3.5-MoE-instruct}{Phi-3~\cite{abdin2024phi}} & 41.9B & 2/16/0 & 32 & 32 & 4096 & 6400 & 6400 & MicroSoft & 2024.08 \\
\hline

\href{https://huggingface.co/microsoft/GRIN-MoE}{GRIN-MoE~\cite{liu2024grin}} & 41.9B & 2/16/0 & 32 & 32 & 4096 & 6400 & 6400 & MicroSoft & 2024.09 \\
\hline

\href{https://huggingface.co/tencent/Tencent-Hunyuan-Large/tree/main/Hunyuan-A52B-Pretrain}{Hunyuan-Large~\cite{sun2024hunyuan}} & 389B & 1/16/1 & 64 & 80 & 6400 & 18304 & 18304 & Tencent & 2024.11 \\
\hline

\href{https://huggingface.co/deepseek-ai/DeepSeek-V3-Base}{DeepSeek-V3~\cite{liu2024deepseek}} & 671B & 8/256/1 & 61 & 128 & 7168 & 18432 & 2048 & DeepSeek-AI & 2024.12 \\
\hline

\href{https://huggingface.co/MiniMaxAI/MiniMax-Text-01}{MiniMax-Text-01~\cite{minimax2025minimax01scalingfoundationmodels}} & 456B & 2/32/0 & 80 & 64 & 6144 & 9216 & 9216 & MiniMax-AI & 2025.1 \\
\hline

\hline

\end{tabular}
\caption{A List of SoTA MoEs. Param. represents the number of total parameters. Experts are listed according to the format of the number of activation experts, total experts, and shared experts. \#L represents the number of hidden layers, \#H represents the number of attention heads. $d_{model}$ is the hidden size, $d_{ffn}$ is the intermediate size of FFNs, $d_{expert}$ is the intermediate size of FFN experts.}\label{tab:sota-moe}
\end{table}

\begin{figure}[t]
    \centering
    
\tikzset{
    basic/.style  = {draw, text width=3cm, align=center, font=\sffamily, rectangle},
    root/.style   = {basic, rounded corners=2pt, thin, align=center, text width=1.8cm, fill=red!20},
    onode/.style = {basic, thin, rounded corners=2pt, align=center, fill=green!20,text width=1.5cm,},
    xnode/.style = {basic, thin, rounded corners=2pt, align=center, fill=blue!20,text width=2cm,},
    wnode/.style = {basic, thin, rounded corners=2pt, align=center, fill=pink!20, text width=2cm},
    tnode/.style = {basic, thin, rounded corners=2pt, align=left, fill=yellow!20, text width=30em},
    ttnode/.style = {basic, thin, rounded corners=2pt, align=left, fill=yellow!20, text width=23em},
    edge from parent/.style={draw=black, edge from parent fork right}

}

\begin{forest} for tree={
    grow'=east,
    growth parent anchor=west,
    parent anchor=east,
    child anchor=west,
    edge path={\noexpand\path[\forestoption{edge},->, >={latex}] 
         (!u.parent anchor) -- +(5pt,0pt) |-  (.child anchor) 
         \forestoption{edge label};},
}
[MoE Inference Optimization, root, l sep=5mm, before typesetting nodes={if n=1{anchor=south}{ };},
    [Model Level, onode,  l sep=5mm,
        [Architecture Design, xnode, l sep=5mm,calign=child edge,
            [MoH~\cite{jin2024moh}\,
            JetMoE~\cite{shen2024jetmoe}\,
            ModuleFormer~\cite{shen2023moduleformer}\,
            DS-MoE~\cite{pan2024dense}\,
            MoA~\cite{zhang2022mixture}\,
            SwitchHead~\cite{csordas2023switchhead}\,
            BAM~\cite{zhang2024bam}\, MAE~\cite{peng-etal-2020-mixture}\, SUT~\cite{tan2023sparse}\, MoEUT~\cite{csordas2024moeut}\,
            MoE++~\cite{jin2024moe++}\, MoELoRA~\cite{luo2024moelora}\,
            Pre-gated MoE~\cite{hwang2024pre}\,
            SCoMoE~\cite{xiongscomoe}\,
            COMET~\cite{ibrahim2023comet},
            ttnode]
        ]
        [Model Compression, xnode, l sep=5mm,calign=child edge,
            [TSEP~\cite{chen2022task}\, NAEE~\cite{lu2024not}\, UNCURL~\cite{sarkar2024revisiting}\, PEMPE~\cite{chowdhury2024provably}\, SEER-MoE~\cite{muzio2024seer}\, DEK~\cite{zhang2024diversifying}\, EEP~\cite{liu2024efficient}\, MC-SMoE~\cite{li2023merge}\, MoE-Pruner~\cite{xie2024moe}\, ModuleFormer~\cite{shen2023moduleformer}\, STUN~\cite{lee2024stun}\, MoE-Compression~\cite{he2024demystifying}\,
            MC-MoE~\cite{huang2024mc}\, MoE-CSP~\cite{kim2022says}\, MoQE~\cite{kim2023mixture}\, QMoE~\cite{frantar2023qmoe}\, CMoE~\cite{yuancompressed}\, MoE-MPTQS~\cite{imani2024mixture}\, HOBBIT~\cite{tang2024hobbit}\, EdgeMoE~\cite{yi2023edgemoe}\, QMoE-Benchmark~\cite{li2024examining}\,
            LLaVA-MoD~\cite{shu2024llava}\,  DeepSpeed-MoE~\cite{DeepSpeedmoe}\, MoE-KD~\cite{salinas2022knowledge}\, OneS~\cite{xue2022one}\, LaDiMo~\cite{kim2024ladimo}\, Switch Transformers~\cite{fedus2022switch}\, ELSM~\cite{artetxe2021efficient}\,
            MPOE~\cite{gao2022parameter}\, MoE-$I^{2}$~\cite{yang2024moe}\,
            FoE~\cite{wang2023fusing}\,
            MEO~\cite{he2023merging}\,
            HC-SMoE~\cite{chen2024retraining}\,
            Park \textit{et al.}~\cite{park2024learning}\,
            Branch-Train-Mix~\cite{sukhbaatar2024branch}\,
            Branch-Train-Merge~\cite{li2022branch}\,
            HyperMoE~\cite{zhao-etal-2024-hypermoe}\,
            LiteMoE~\cite{zhuang2024litemoe},
            ttnode]
        ]
        [Algorithm Improvement, xnode,  l sep=5mm,calign=child edge,
            [Li \textit{et al.}~\cite{li2023adaptive}\,
            AdaptMoE~\cite{zhong2024adapmoe}\,
            DynMoE~\cite{guo2024dynamic}\,
            XMoE~\cite{yang2024xmoe}\,
            XFT~\cite{ding2024xft}\,
            Switch Transformers~\cite{fedus2022switch}\,ELSM~\cite{artetxe2021efficient}\,
            OneS~\cite{xue2022one}\,
            TSEP~\cite{chen2022task}\,
            EWA~\cite{huang2023experts}\,
            DA-MoE~\cite{aghdam2024damoedynamicexpertallocation}\,
            AdaMoLE~\cite{liu2024adamolefinetuninglargelanguage},
            ttnode]
        ]
    ] 
    [System Level, onode,  l sep=5mm,
        [Expert Parallel, xnode, l sep=5mm,calign=child edge,
            [Gshard~\cite{lepikhin2020gshard}\, FastMoE~\cite{he2021fastmoe}\, Tutel~\cite{hwang2023tutel}\, MoESys~\cite{10528887}\, Alpa~\cite{Alpa}\, BaGuaLu~\cite{BaGuaLu}\, SmartMoE~\cite{288691}\, 
            SwitchTransformers~\cite{fedus2022switch}\, HashLayer~\cite{roller2021hash}\, Prophet~\cite{10319949}\, MoE-Prediction~\cite{cong2024prediction}\, Lazarus~\cite{wu2024lazarus}\, FlexMoE~\cite{nie2023flexmoe}\, MoE-Deploy~\cite{huang2023towards}\, Brainstorm~\cite{Brainstorm}\, Lynx~\cite{gupta2024lynx}\, BaseLayers~\cite{lewis2021base}\, MoE-ECR~\cite{moeecr}\,
            Janus~\cite{10.1145/3603269.3604869}\, HetuMoE~\cite{nie2022hetumoe}\, DeepSpeed-MoE~\cite{DeepSpeedmoe}\, DeepSpeed-TED~\cite{DeepSpeedted}\, Lina~\cite{288705}\, ExFlow~\cite{yao2024exploiting}\, TA-MoE~\cite{chen2022ta}\, Aurora~\cite{li2024optimizing}\, LocMoE~\cite{li2024locmoe}\, Parm~\cite{10621327}\,
            ScMoE~\cite{cai2024shortcut}\, HiDup~\cite{HiDup}\, ScheMoE~\cite{10.1145/3627703.3650083}\, PipeMoE~\cite{10228874}\, EPS-MoE~\cite{qian2024eps}\, MoE-SLC~\cite{liu2025optimizing}, 
            ttnode]
        ]
        [Expert Offloading, xnode, l sep=5mm,calign=child edge,
            [HOBBIT~\cite{tang2024hobbit}\, ExpertFlow~\cite{he2024expertflow}\,  SiDA~\cite{du2024sida}\, MoE-Infinity~\cite{xue2024moe}\, Pre-gated MoE~\cite{hwang2024pre}\, Mixtral-Offloading~\cite{eliseev2023fast}\, EdgeMoE~\cite{yi2023edgemoe} DyNN-Offload~\cite{dynnoffload}\, Read-ME~\cite{readmoe}\, ProMoE\cite{song2024promoe}\, AdapMoE~\cite{zhong2024adapmoe}\, MoE-Fiddler~\cite{kamahori2024fiddler}\, EIO-MoE~\cite{yuan2024efficient}\, MoE-Deploy~\cite{huang2023towards}\,  SwapMoE~\cite{kong2023serving}\,   MoE-Lightning~\cite{cao2024moelight}\, CacheMoE~\cite{skliar2024mixture},
            ttnode]
        ] 
    ]
    [Hardware Level, onode,  l sep=5mm,calign=child edge
        [MoNDE~\cite{Kim2024monde}\, FLAME~\cite{Lin2024flame}\, Duplex~\cite{Yun2024Duplex}\, M\textsuperscript{3}ViT~\cite{Liang2022m3vit}\, Edge-MoE~\cite{Sarkar2024edgemoe}\, Space-Mate~\cite{Park2024spacemate}, tnode]
    ]
]
\end{forest}
    \caption{Taxonomy of MoE inference optimization.}
    \label{fig:lit_surv}
\end{figure}
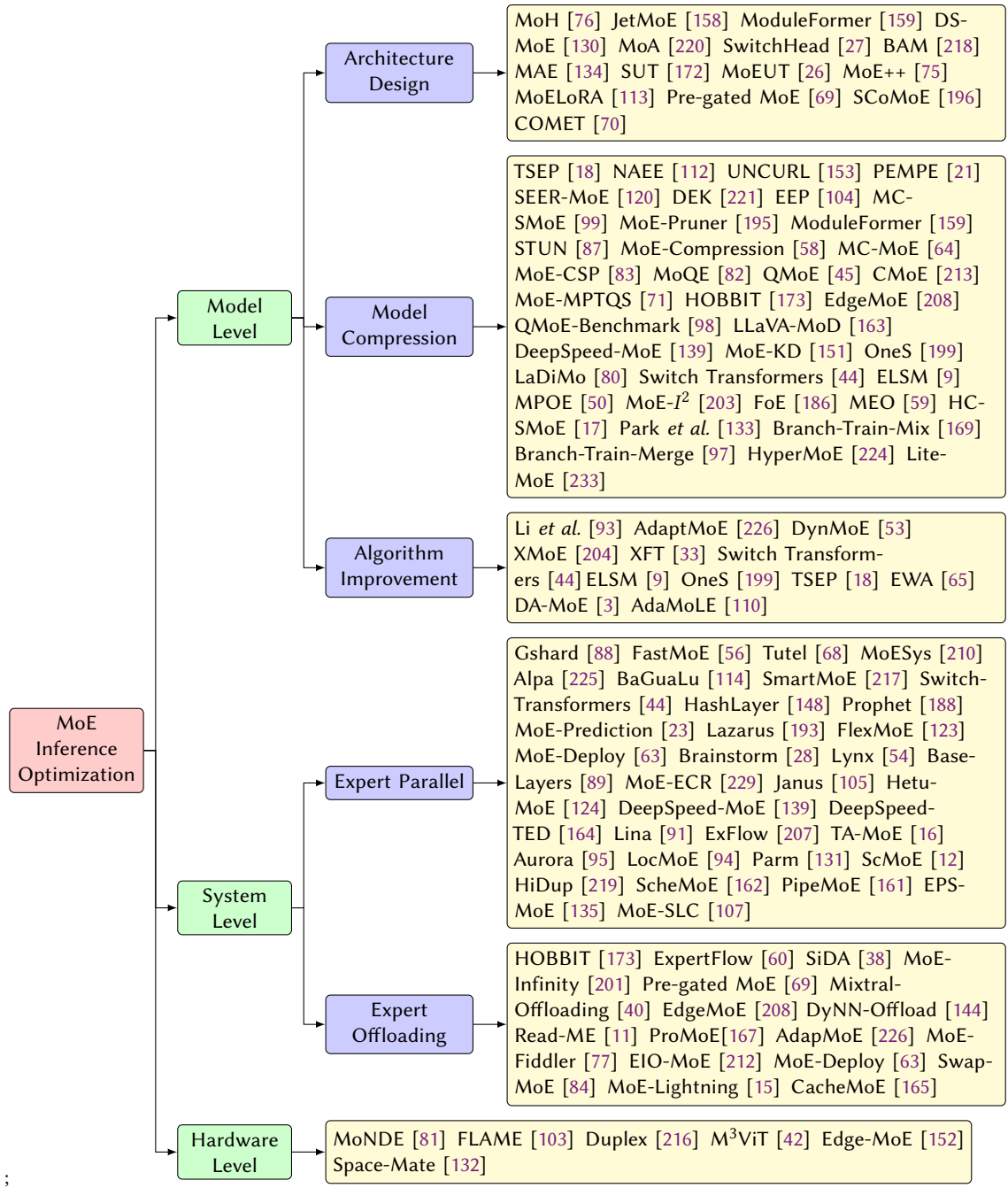

The success of MoE in scaling models has led to its adoption in various state-of-the-art systems. For example, Google's GLaM~\cite{du2022glam} outperforms GPT-3 while using significantly fewer computational resources during inference. Similarly, Mixtral 8x7B~\cite{jiang2024mixtral} has demonstrated competitive performance compared to much larger dense models, while maintaining efficient inference characteristics. 
DeepSeek-V3~\cite{liu2024deepseek}, a recent open-source MoE model, 
has surpassed other open-source alternatives and demonstrated performance comparable to prominent closed-source models such as GPT4-o and Claude-3.5-Sonnet.
Table~\ref{tab:sota-moe} summarizes recent state-of-the-art open-source MoE models that have garnered significant attention, showcasing their rapid evolution and widespread adoption across major tech companies and research institutions. The models range in size from 6.7B to 671B parameters, with varying architectures characterized by different numbers of experts, hidden layers, and attention heads. The chronological progression from 2022 to late 2024 demonstrates increasing model sizes and architectural sophistication, with recent models like DeepSeek-V3 pushing the boundaries in terms of parameter count and model performance. This trend further highlights the strong potential of the MoE architecture in advancing the field of large language models.
These successes have sparked widespread interest in MoE across both academia and industry, leading to innovations in model design~\cite{chowdhury2023patch, wang2020deep, zhang2022mixture}, training techniques~\cite{liu2023moelora, dou2024loramoe, gao2024higher}, and deployment strategies~\cite{chen2022ta, cao2024moelight, 10528887}.

However, the efficient deployment of MoE models for inference presents unique and significant challenges~\cite{tang2024hobbit, yi2023edgemoe, zhong2024adapmoe, hwang2024pre}. The dynamic nature of expert activation patterns introduces complexity in resource management and scheduling that is not present in traditional dense models. 
At the model level, the design of efficient expert architectures faces challenges in balancing model capacity with computational efficiency, while routing mechanisms must optimize expert selection and load distribution. 
The system-level challenges are particularly complex: distributed computation requires sophisticated scheduling algorithms to manage expert placement and activation, load balancing must handle dynamic workload variations across experts, and memory management needs to efficiently handle the loading and unloading of expert parameters. 
Hardware-level challenges stem from the fundamental mismatch between traditional hardware architectures optimized for dense computation and the sparse, dynamic nature of MoE inference. This necessitates specialized acceleration techniques to handle sparse computation patterns, manage memory bandwidth efficiently, and provide flexible computation capabilities for dynamic expert switching. Communication overhead in distributed settings presents another significant challenge, particularly when experts are distributed across different devices or nodes, requiring careful optimization of data movement and synchronization.

Numerous methods have been developed to address these challenges in MoE deployment and inference~\cite{jin2024moh, xie2024moe, DeepSpeedmoe, Sarkar2024edgemoe}. While the rapid growth of research in this field demonstrates its importance, it can also make it difficult to identify key trends and best practices. A critical gap in the existing literature is the absence of a systematic framework for analyzing and developing comprehensive inference optimization solutions for MoE models.
To bridge this gap, our work offers a comprehensive survey of inference optimization techniques for MoE models. We propose a taxonomical framework that categorizes optimization approaches into model-level, system-level, and hardware-level optimizations, as illustrated in Figure \ref{fig:lit_surv}. This framework provides a structured approach to understanding and comparing different optimization techniques. While there are related surveys on LLM efficiency~\cite{yuan2024llm, li2024llm, li2024large, treviso2023efficient, wan2023efficient, bai2024beyond, zhou2024survey, xu2024survey} and MoE architectures~\cite{202408.0583, cai2024survey, fedus2022review}, our work is the first to specifically focus on inference optimization techniques for MoE models. We systematically analyze optimization approaches at different abstraction levels, from model architecture to hardware acceleration, providing a valuable resource for researchers and practitioners working deploy MoE models for different real-world applications.

The remainder of this survey is organized as follows: Section \ref{sec:background} provides background on MoE models and their inference characteristics. Sections 3-5 detail optimization techniques at the model, system, and hardware levels respectively. Section \ref{sec:future} discusses future challenges and opportunities, and Section \ref{sec:conclusion} concludes the survey.

\begin{figure}[t]
    \centering
    \subfloat[Dense Layer]{\includegraphics[width=0.177\linewidth]{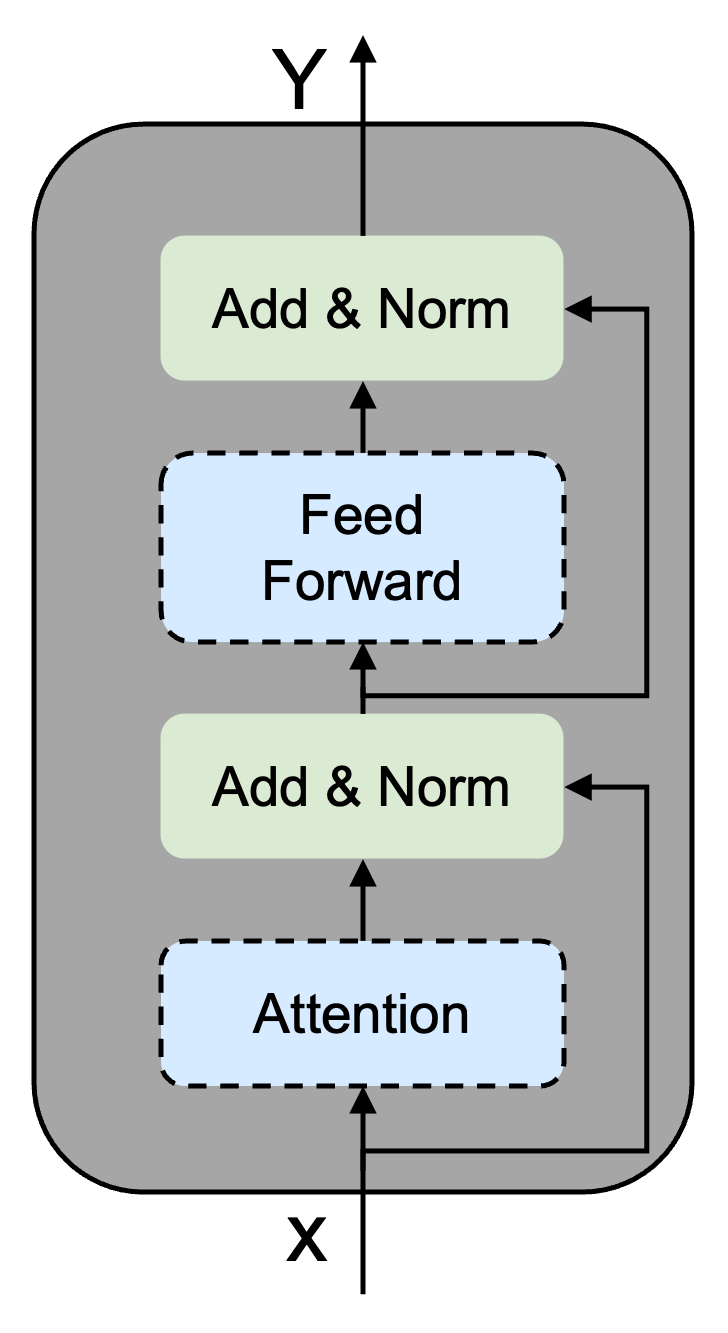}}
    \subfloat[MoE FFN Layer]{\includegraphics[width=0.411\linewidth]{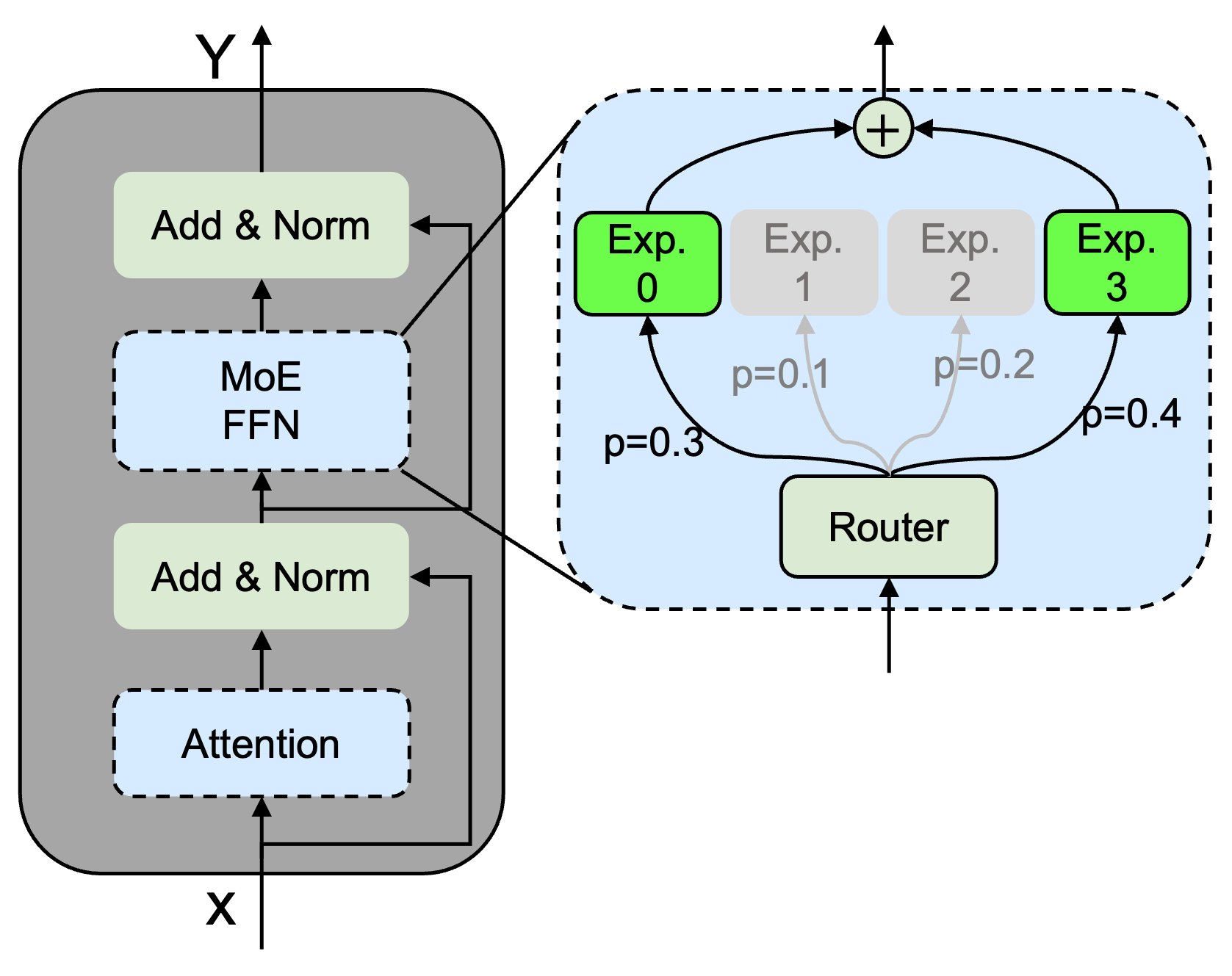}}
    \subfloat[MoE Attention+FFN Layer]{\includegraphics[width=0.411\linewidth]{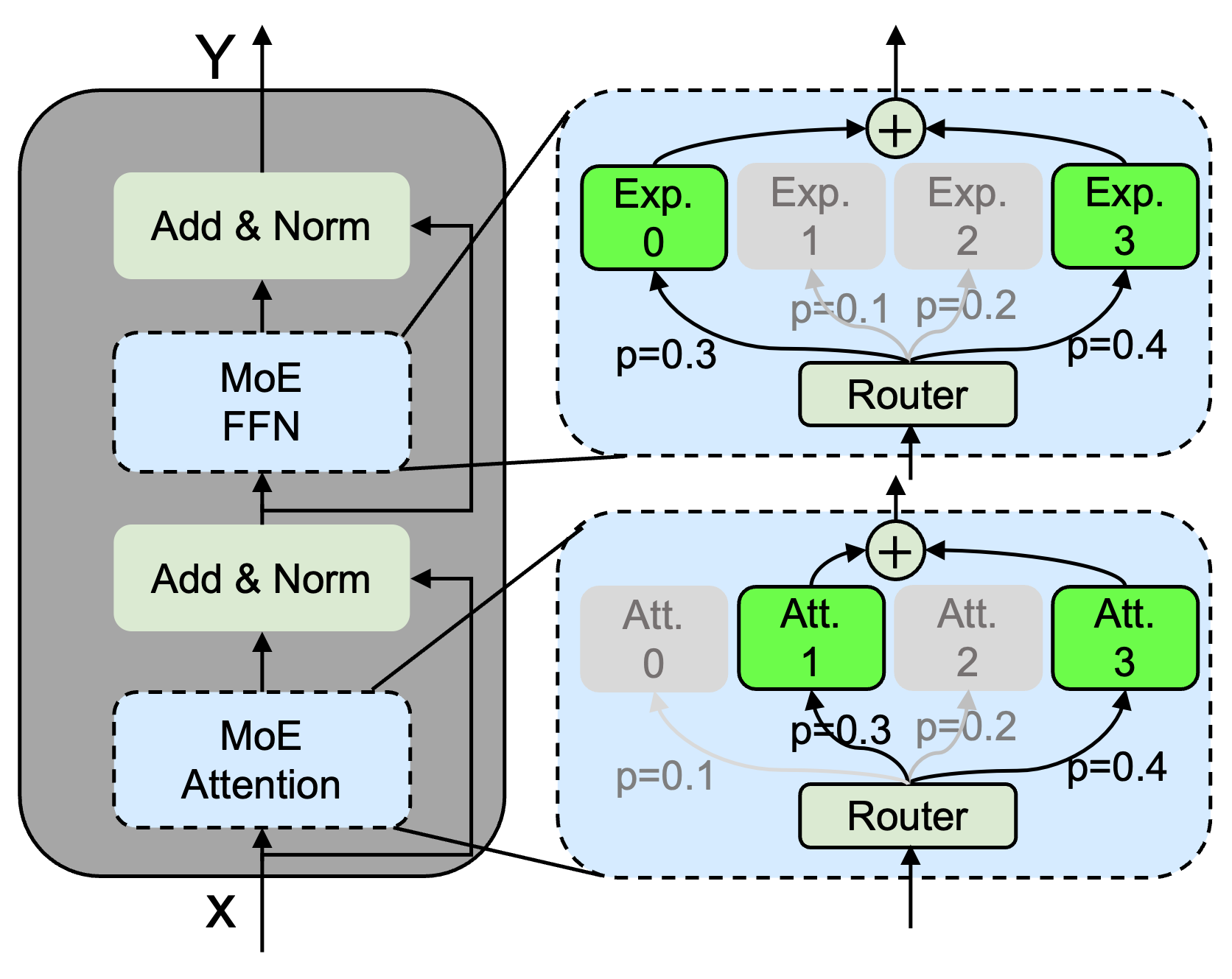}}
    \caption{
    Architectural comparison of dense layer with MoE layers: (a) conventional dense transformer layer, (b) transformer layer with MoE-based feed-forward network, and (c) transformer layer with both MoE-based attention and feed-forward networks. 
    }
  \label{fig:moe implementation}
\end{figure}

\section{Fundamentals of Mixture of Experts}\label{sec:background}

 MoE represents a significant architectural paradigm in neural networks, particularly in large language models, where it enables conditional computation through sparse activation mechanisms~\cite{cai2024survey}. At its core, an MoE architecture consists of a routing network $R(x)$ and a set of expert networks ${E_1, E_2, ..., E_N}$, where $N$ denotes the total number of experts. The fundamental principle of MoE can be expressed as $y = \sum_{i=1}^{N} g_i(x) \cdot E_i(x)$, where $g_i(x)$ represents the gating function for expert $i$, and $E_i(x)$ is the output of expert $i$.

As illustrated in Figure~\ref{fig:moe implementation}, existing studies typically use the MoE module to replace part of the traditional dense layer, thereby forming a sparse MoE layer. While most research focuses on substituting the FFN module with the MoE module~\cite{jiang2024mixtral, deepseekai2024deepseekv2strongeconomicalefficient, qwen_moe, abdin2024phi, jin2024moe++}, some have also explored replacing the Attention module~\cite{shen2024jetmoe, jin2024moh, shen2023moduleformer, zhang2022mixture}. 

The typical MoE model inference procedure follows a sequence of expert selection, parallel computation, and output aggregation. 
First, the router computes expert selection probabilities:

\begin{equation}
\label{eq:routing_prob}
\theta = \text{Softmax}(R(x))
\end{equation}
where $x \in \mathbb{R}^d$ is the input token embedding, $R(\cdot)$ is the routing function, and $\theta \in \mathbb{R}^N$ represents the expert selection probabilities for $N$ total experts.
Next, the top-k experts are selected based on these probabilities:

\begin{equation}
\label{eq:expert_selection}
E_{\text{selected}} = \text{TopK}(\theta, K)
\end{equation}
where $E_{\text{selected}}$ contains the indices of the K selected experts, $K \leq N$.
The selected experts then process the input in parallel:

\begin{equation}
\label{eq:expert_computation}
y_i = E_i(x), \quad \forall i \in E_{\text{selected}}
\end{equation}
where $E_i(\cdot)$ represents the computation of expert $i$, and $y_i$ is its output.
Finally, the expert outputs are combined through weighted aggregation:
\begin{equation}
\label{eq:output_aggregation}
y = \sum_{i \in E_{\text{selected}}} \frac{\theta_i}{\sum_{j \in E_{\text{selected}}} \theta_j} \cdot y_i
\end{equation}

The MoE architecture offers several compelling advantages that have contributed to its growing adoption in modern neural networks. First, the conditional computation mechanism enables significant computational savings compared to dense models of similar capacity. By activating only a subset of experts for each input token, MoE models can process information more efficiently while maintaining high performance levels. This is particularly valuable in resource-constrained environments or when scaling to larger model sizes.

Second, the specialization of individual experts allows for more nuanced and accurate processing of different input patterns. Each expert can develop specialized knowledge for specific aspects of the input space, leading to more refined and targeted computations. This specialization is particularly beneficial in language models, where different experts can focus on various linguistic patterns, domains, or tasks.

Third,  the dynamic routing mechanism enables adaptive computation based on input complexity. More challenging or nuanced inputs can engage multiple experts with complementary specializations, while simpler inputs might require only a small subset of experts. This adaptive resource allocation helps optimize the computation-performance trade-off and potentially improves both efficiency and effectiveness.

By partitioning dense models into relatively independent expert models and dynamically activating specific subsets (or the entire set) of experts based on each input token, the model's overall performance can be significantly enhanced with only a marginal increase in inference computation. This approach clearly demonstrates the MoE model's exceptional flexibility and scalability.

\section{Model-level Optimizations}\label{sec:model}

\begin{figure}
    \centering
    
\tikzset{
    basic/.style  = {draw, text width=3cm, align=center, font=\sffamily, rectangle},
    root/.style   = {basic, rounded corners=2pt, thin, align=center, text width=1.5cm, fill=red!20},
    onode/.style = {basic, thin, rounded corners=2pt, align=center, fill=green!20,text width=1.5cm,},
    xnode/.style = {basic, thin, rounded corners=2pt, align=center, fill=blue!20,text width=2cm,},
    wnode/.style = {basic, thin, rounded corners=2pt, align=center, fill=pink!20, text width=2cm},
    tnode/.style = {basic, thin, rounded corners=2pt, align=left, fill=yellow!20, text width=29em},
    ttnode/.style = {basic, thin, rounded corners=2pt, align=left, fill=yellow!20, text width=22em},
    edge from parent/.style={draw=black, edge from parent fork right}

}

\begin{forest} for tree={
    grow'=east,
    growth parent anchor=west,
    parent anchor=east,
    child anchor=west,
    edge path={\noexpand\path[\forestoption{edge},->, >={latex}] 
         (!u.parent anchor) -- +(5pt,0pt) |-  (.child anchor) 
         \forestoption{edge label};},
}
[Model Level, onode,  l sep=5mm,  before typesetting nodes={if n=1{anchor=south}{ };},
    [Architecture Design, xnode, l sep=5mm,
        [MoE-based Attention, wnode, l sep=5mm,calign=child edge,
            [MoH~\cite{jin2024moh}\,
            JetMoE~\cite{shen2024jetmoe}\,
            ModuleFormer~\cite{shen2023moduleformer}\,
            DS-MoE~\cite{pan2024dense}\,
            MoA~\cite{zhang2022mixture}\,
            SwitchHead~\cite{csordas2023switchhead}\,
            BAM~\cite{zhang2024bam}\, MAE~\cite{peng-etal-2020-mixture}\, SUT~\cite{tan2023sparse}\, MoEUT~\cite{csordas2024moeut},
            ttnode]
        ]
        [MoE-based FFN, wnode, l sep=5mm,calign=child edge,
            [MoE++~\cite{jin2024moe++}\, MoELoRA~\cite{luo2024moelora}\,
            Pre-gated MoE~\cite{hwang2024pre}\,
            SCoMoE~\cite{xiongscomoe}\,
            COMET~\cite{ibrahim2023comet},
            ttnode]
        ]
    ]
    [Model Compression, xnode, l sep=5mm,
        [Expert \quad Pruning, wnode,l sep=5mm,calign=child edge,
            [TSEP~\cite{chen2022task}\, NAEE~\cite{lu2024not}\, UNCURL~\cite{sarkar2024revisiting}\, PEMPE~\cite{chowdhury2024provably}\, SEER-MoE~\cite{muzio2024seer}\, MoE-$I^{2}$~\cite{yang2024moe}\, DEK~\cite{zhang2024diversifying}\, EEP~\cite{liu2024efficient}\, MC-SMoE~\cite{li2023merge}\, MoE-Pruner~\cite{xie2024moe}\, ModuleFormer~\cite{shen2023moduleformer}\, STUN~\cite{lee2024stun}\, MoE-Compression~\cite{he2024demystifying}\,
            FoE~\cite{wang2023fusing}\,
            MEO~\cite{he2023merging}\,
            HC-SMoE~\cite{chen2024retraining}\,
            Park \textit{et al.}~\cite{park2024learning}\,
            Branch-Train-Mix~\cite{sukhbaatar2024branch}\,
            Branch-Train-Merge~\cite{li2022branch}\,
            HyperMoE~\cite{zhao-etal-2024-hypermoe}\,
            LiteMoE~\cite{zhuang2024litemoe},
            ttnode]
        ]
        [Expert Quantization, wnode,l sep=5mm,calign=child edge,
            [MC-MoE~\cite{huang2024mc}\, MoE-CSP~\cite{kim2022says}\, MoQE~\cite{kim2023mixture}\, QMoE~\cite{frantar2023qmoe}\, CMoE~\cite{yuancompressed}\, MoE-MPTQS~\cite{imani2024mixture}\, HOBBIT~\cite{tang2024hobbit}\, EdgeMoE~\cite{yi2023edgemoe}\, QMoE-Benchmark~\cite{li2024examining},
            ttnode]
        ]
        [Expert Distillation, wnode,l sep=5mm,calign=child edge,
            [LLaVA-MoD~\cite{shu2024llava}\,  DeepSpeed-MoE~\cite{DeepSpeedmoe}\, MoE-KD~\cite{salinas2022knowledge}\, OneS~\cite{xue2022one}\, LaDiMo~\cite{kim2024ladimo}\, CMoE~\cite{yuancompressed}\, Switch Transformers~\cite{fedus2022switch}\, ELSM~\cite{artetxe2021efficient}, ttnode]   
        ]
        [Expert Decomposition, wnode, l sep=5mm,calign=child edge,
            [MPOE~\cite{gao2022parameter}\, SMoE~\cite{li2023merge}\, MoE-$I^{2}$~\cite{yang2024moe}, ttnode]
        ]
    ]
    [Algorithm Improvement, xnode,  l sep=5mm,
        [Dynamic Gating, wnode, l sep=5mm,calign=child edge,
            [Li \textit{et al.}~\cite{li2023adaptive}\,
            AdaptMoE~\cite{zhong2024adapmoe}\,
            DynMoE~\cite{guo2024dynamic}\,
            XMoE~\cite{yang2024xmoe}\,
            DA-MoE~\cite{aghdam2024damoedynamicexpertallocation},
            ttnode]
        ]
        [Sparse to Dense, wnode, l sep=5mm,calign=child edge,
            [XFT~\cite{ding2024xft}\,
            Switch Transformers~\cite{fedus2022switch}\,
            ELSM~\cite{artetxe2021efficient}\,
            OneS~\cite{xue2022one}\,
            TSEP~\cite{chen2022task}\,
            EWA~\cite{huang2023experts}\,
            AdaMoLE~\cite{liu2024adamolefinetuninglargelanguage},
            ttnode]
        ] 
    ] 
]
\end{forest}
    \caption{Model-level inference optimization techniques for MoE.}
    \label{fig:model-opt}
\end{figure}
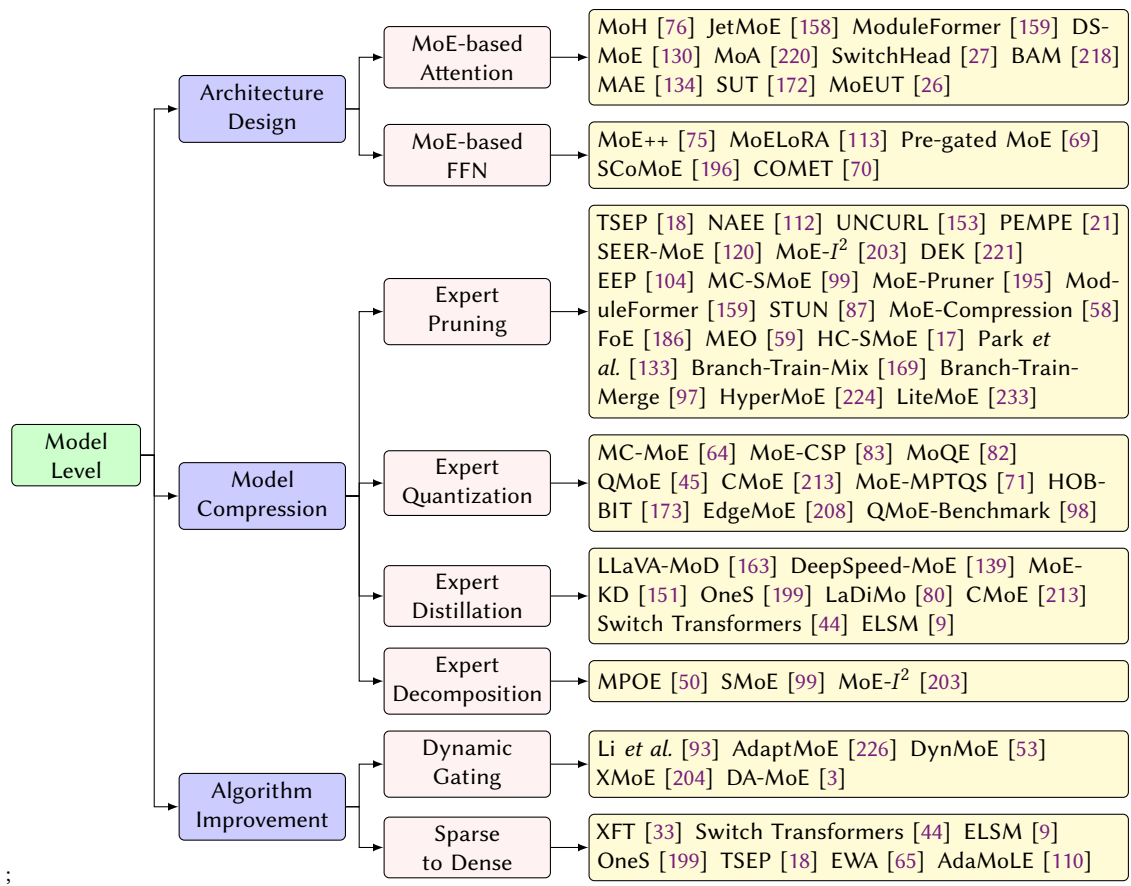

Model-level optimizations aim to enhance the inherent structure and efficiency of MoE models through systematic improvements in architecture, parameter optimization, and algorithmic design. These optimizations can be broadly categorized into three main areas: efficient model architecture design, model compression techniques, and algorithmic improvements. Architecture design focuses on developing more efficient expert and attention structures, while compression techniques aim to reduce model size and memory footprint through methods such as pruning, quantization, and knowledge distillation. Algorithmic improvements concentrate on enhancing the dynamic aspects of MoE models, including routing mechanisms and expert combination strategies. Figure~\ref{fig:model-opt} illustrates the detailed taxonomy of model-level optimization that is described in this section.

\subsection{Efficient Model Architecture Design}

A transformer block typically consists of two main components: the attention module and the 
FFN module. To build better MoE models, many studies focus on designing improved versions of both the attention and FFN modules, aiming to achieve strong performance while maintaining high efficiency.

\subsubsection{MoE-based Attention Design}

In addition to the typical application of the MoE structure in the FFN  module of the transformer layer, current studies explore how to incorporate MoE into the Attention module for improved performance. MAE~\cite{peng-etal-2020-mixture} was the first to explain the multi-head attention mechanism from the MoE perspective, using a learned gating function to activate different experts for different inputs, with each expert consisting of \(n-1\) heads. To further optimize MoE-based attention modules, existing research proposes various structures. MoA~\cite{zhang2022mixture} and BAM~\cite{zhang2024bam} select \(k\) heads for a given input and share key projection and value projection weights among all heads, while SwitchHead~\cite{csordas2023switchhead} shares key projection and query projection weights to enhance computational efficiency. MoH~\cite{jin2024moh} introduces shared heads and a two-stage routing process to further improve the standard MoE method, offering an advantage over MoA. Building upon MoA, ModuleFormer~\cite{shen2023moduleformer} extends sparse modules to both the attention and feed-forward layers, allowing for the easy addition and removal of modules. Inspired by MoA and ModuleFormer, JetMoE-8B~\cite{shen2024jetmoe} develops a powerful open-source model featuring sparse attention and sparse feed-forward layers, while DS-MoE~\cite{pan2024dense} proposes a hybrid dense training and sparse inference framework for efficient training and inference. Additionally, SUT~\cite{tan2023sparse} and MoEUT~\cite{csordas2024moeut} use sparse attention and sparse feed-forward layers to construct the efficient Sparse Universal Transformer model, which shares parameters across all layers.

\subsubsection{MoE-based FFN Design}

To enhance the efficiency of MoE-based models, current research explores various variants of the standard MoE module. MoE++~\cite{jin2024moe++} introduces three types of zero-computation experts based on standard experts, aimed at reducing computational overhead. SCoMoE~\cite{xiongscomoe} leverages a structured all-to-all communication approach, inspired by hierarchical communication topologies, to reduce communication costs during parallel MoE computation. Pre-gated MoE~\cite{hwang2024pre} proposes a pre-gated MoE module that prefetches the required experts to improve inference speed on memory-constrained devices. COMET~\cite{ibrahim2023comet} introduces a tree-based sparse expert selection mechanism to optimize the traditional gating module, which typically relies on a linear approach. Additionally, MoELoRA~\cite{luo2024moelora} reimagines LoRA as a MoE for more parameter-efficient fine-tuning.

\subsection{Model Compression Techniques}

Model compression is a popular area of research for reducing model size in current LLM studies, with techniques such as pruning~\cite{men2024shortgpt, li2023losparse}, quantization~\cite{wang2023bitnet, frantar2022gptq}, knowledge distillation~\cite{gu2023knowledge, agarwal2023gkd}, and low-rank decomposition~\cite{wang2024svd, yuan2023asvd}. Since experts constitute the majority of the weights in MoE models (e.g., 96\% for Mixtral-8x7B~\cite{jiang2024mixtral}), most MoE-related compression efforts focus on applying these common techniques specifically to the experts.

\begin{figure}[t]
    \centering
    \subfloat[Expert Pruning]{\includegraphics[width=0.5\linewidth]{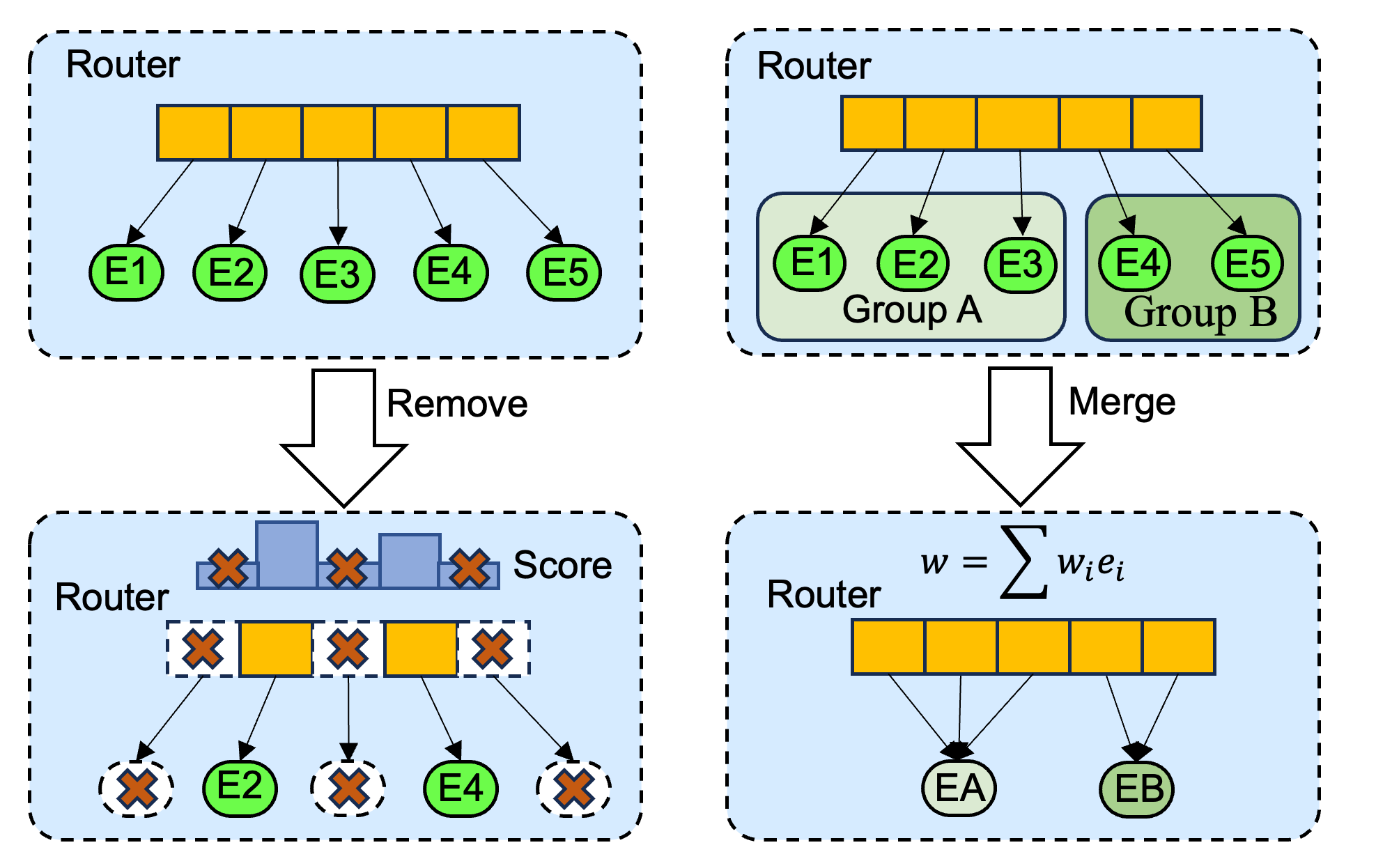}}
    \subfloat[Expert Quantization]{\includegraphics[width=0.5\linewidth]{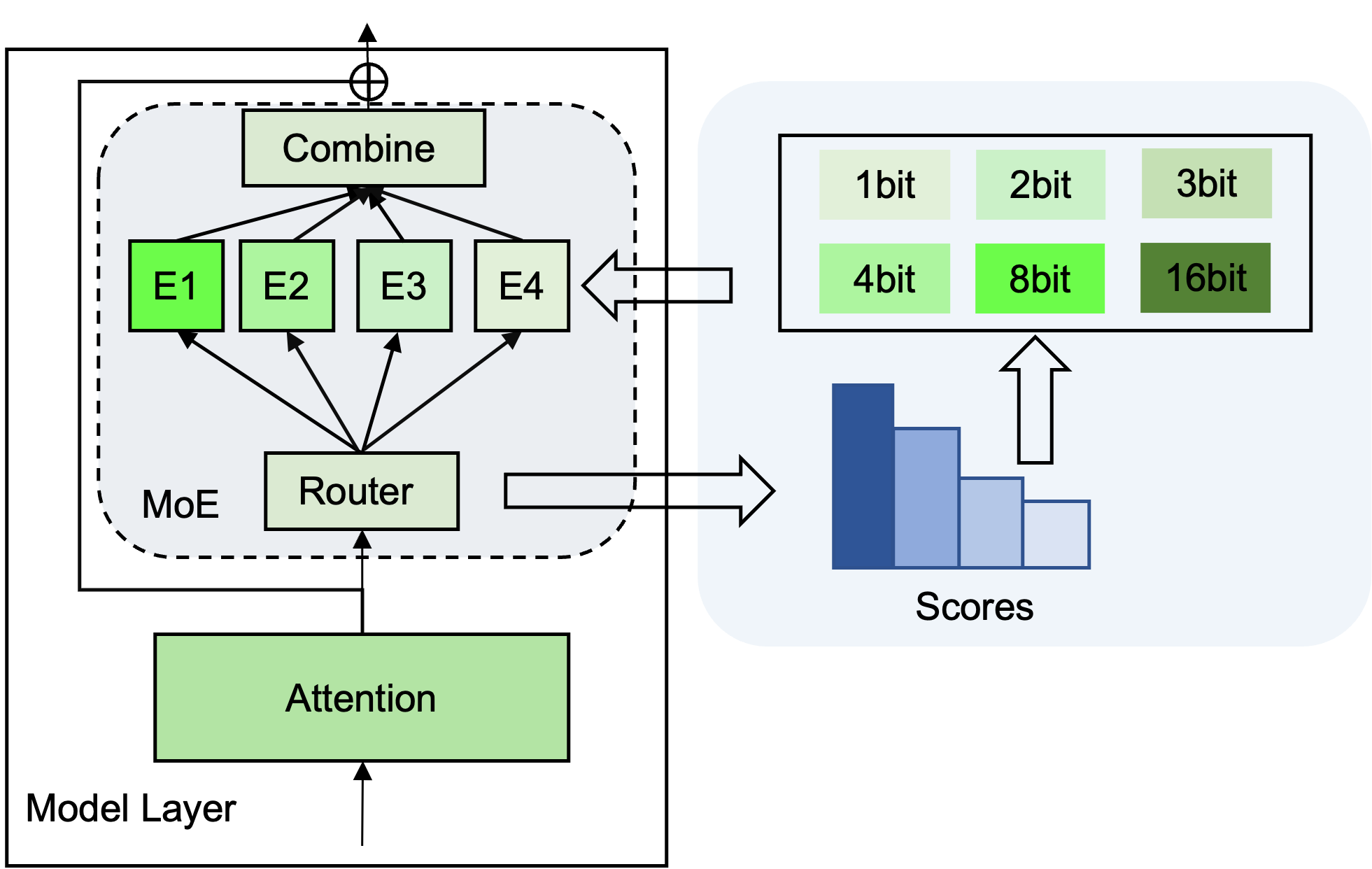}}\\
    \subfloat[Expert Distillation]{\includegraphics[width=0.5\linewidth]{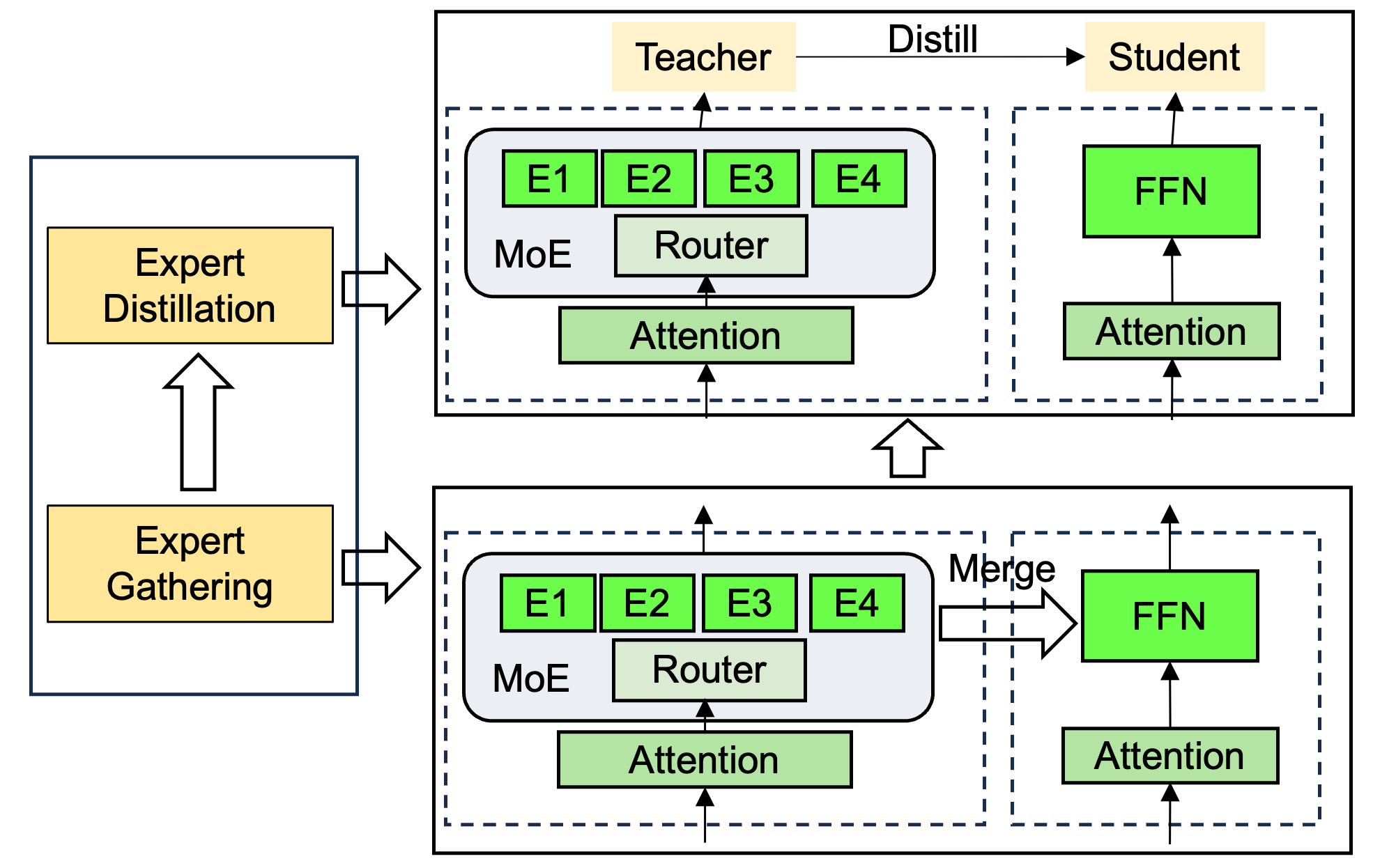}}
    \subfloat[Expert Decomposition]{\includegraphics[width=0.5\linewidth]{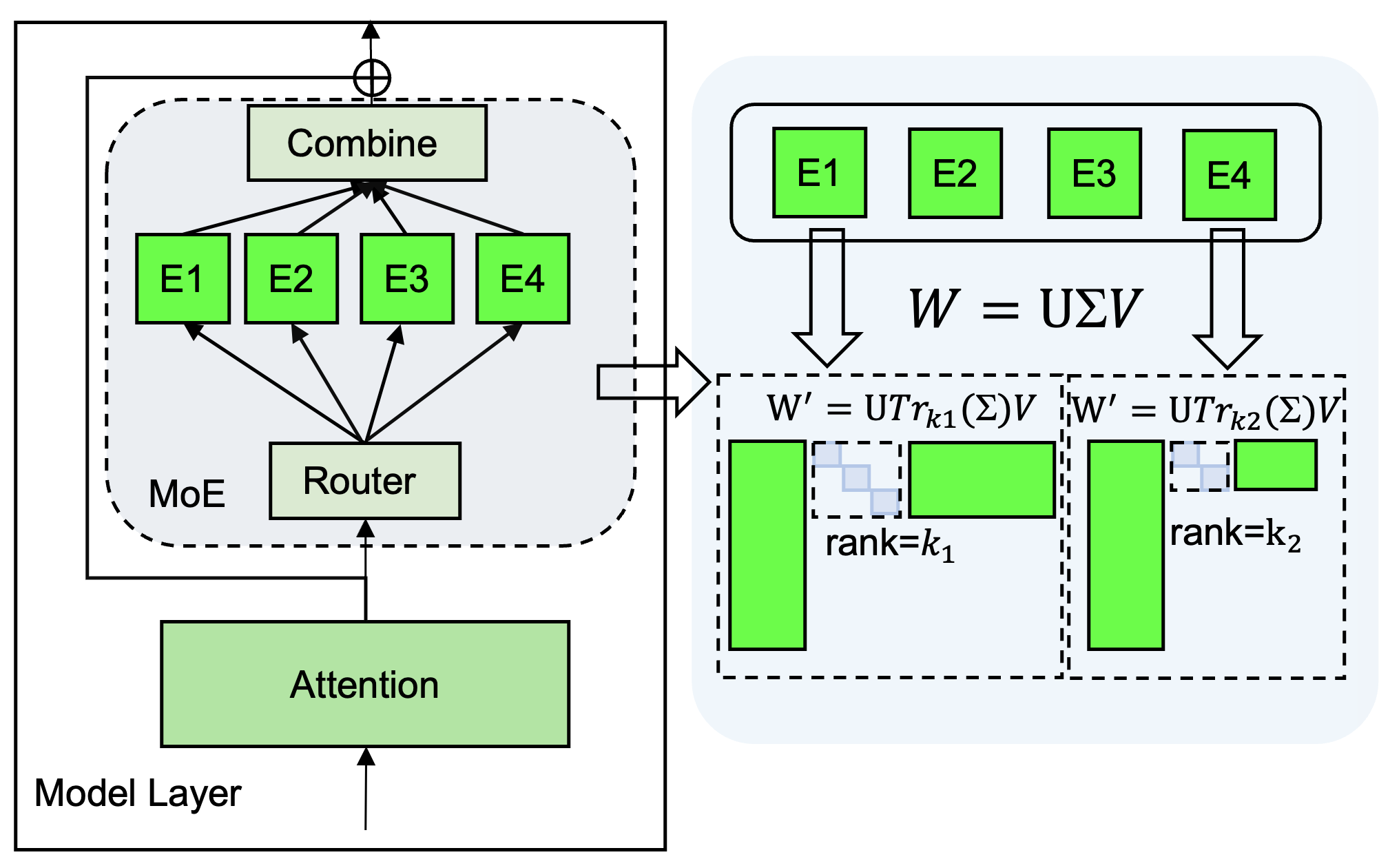}}
    \caption{Compression techniques for MoE models.}
  \label{fig:compression}
\end{figure}

\subsubsection{Expert Pruning}

\begin{table}[t]
\centering
\small
\begin{tabular}{c|c|c|c|cc|c}
\hline
& \multicolumn{1}{c|}{\multirow{2}{*}{Sparsity}} 
& \multicolumn{1}{c|}{\multirow{2}{*}{TSA}}
& \multicolumn{1}{c|}{\multirow{2}{*}{Datasets of Fine-tuning}}
& \multicolumn{2}{c|}{Structured} 
& \multirow{2}{*}{Unstructured} \\ \cline{5-6}
& & & & Delete & Merge & \\ \hline 
TSEP~\cite{chen2022task}              & 96.875\%                                      & S                                       & GLUE~\cite{wang2018glue} SQuAD~\cite{rajpurkar2016squad}                                              & \ding{51} &                        &                                     \\
NAEE~\cite{lu2024not}              & 50\%                                          & S                                       & MetaMathQA~\cite{yu2023metamath}                                              & \ding{51} &                        &                                     \\
UNCURL~\cite{sarkar2024revisiting}            & 75\%                                          & S                                       & FLAN~\cite{longpre2023flan}                                                    & \ding{51} &                        &                                     \\
PEMPE~\cite{chowdhury2024provably}             & 75\%                                          & A                                       & CIFAR-10 CIFAR-100~\cite{krizhevsky2009learning} ImageNet~\cite{russakovsky2015imagenet}                             & \ding{51} &                        &                                     \\
SEER-MoE~\cite{muzio2024seer}          & 25\%                                          & S                                       & MMLU~\cite{hendrycks2020measuring} SST5~\cite{socher2013recursive}                                               & \ding{51} &                        &                                     \\
MoE-$I^{2}$~\cite{yang2024moe}       & 53.98\%                                       & A                                       & Alpaca~\cite{taori2023stanford}                                                  & \ding{51} &                        &                                     \\
DEK~\cite{zhang2024diversifying}               & 75\%                                          & S                                       & C4~\cite{raffel2020exploring}                                                      &                        & \ding{51} &                                     \\
EEP~\cite{liu2024efficient}               & 75\%                                          & A                                       & NA                                                        & \ding{51} & \ding{51} &                                     \\
MC-SMoE~\cite{li2023merge}           & 75\%                                          & S                                       & eight datasets &                        & \ding{51} &                                     \\
MoE-Pruner~\cite{xie2024moe}        & 50\%                                          & A                                       & C4~\cite{raffel2020exploring}                                                      &                        &                        & \ding{51}              \\
STUN~\cite{lee2024stun}              & 70\%                                          & A                                       & C4~\cite{raffel2020exploring}                                                      &                        &                        & \ding{51}              \\
MoE-compression~\cite{he2024demystifying}   & 50\%                                          & A                                       & Lima~\cite{zhou2024lima} MetaMathQA~\cite{yu2023metamath}                                         & \ding{51} &                        & \ding{51}             \\ \hline
\end{tabular}
\caption{A comparison of pruning methods. Sparsity indicates the max ratio of pruning. TSA represents task specific or agnostic. EEP~\cite{liu2024efficient} does not require additional fine-tuning. Finetuning datasets of MS-SMoE~\cite{li2023merge} are SST2~\cite{socher2013recursive} MRPC~\cite{dolan2005automatically} MultiRC~\cite{khashabi2018looking} COPA~\cite{roemmele2011choice} WinoGrande~\cite{sakaguchi2021winogrande} SQuAD~\cite{rajpurkar2016squad} WikiQA~\cite{yang2015wikiqa}and HotpotQA~\cite{yang2018hotpotqa}.}
\label{tab:pruning}
\end{table}

Due to the large number of parameters in MoE-based models, current research explores pruning methods to reduce the number of parameters in MoE experts. These methods are generally divided into structured and unstructured pruning. Most studies focus on structured expert pruning, which aims to reduce the number of experts in the MoE layer while maintaining model accuracy. As shown in Figure~\ref{fig:compression}-(a), some approaches directly remove unimportant experts (left side), while others merge groups of experts into a single expert (right side).  TSEP~\cite{chen2022task} removes non-professional experts for the target downstream task while fine-tuning professional experts to preserve model performance. They also demonstrate the superiority of the eager expert pruning paradigm over other possible solutions like two-pass optimization or staged expert pruning. NAEE~\cite{lu2024not} eliminates unimportant experts by evaluating expert combinations on a small calibration dataset to minimize accuracy loss, simultaneously reduce model sizes and increase inference speed. while UNCURL~\cite{sarkar2024revisiting} reduces the number of experts based on MoE router logits. Some insights useful to model design choices considering task-specific inference optimization can give guidance for later stages. PEMPE~\cite{chowdhury2024provably} prunes experts that have smaller changes in the router's $l_{2}$ norm between the pre-trained and fine-tuned models, and will conduct experiments with other compression techniques on SOTA vision MoEs. SEER-MoE~\cite{muzio2024seer} employs a heavy-hitters counting method for expert pruning, then with regularization-based fine-tuning reaches further expert pruning.  MoE-$I^{2}$~\cite{yang2024moe} proposes the Layer-wise Genetic Search and Block-wise KT-Reception Field with the non-uniform pruning ratio to prune unimportant experts. 

In addition to direct pruning, some studies utilize expert merging to reduce the number of experts. For unstructured pruning, MoE-Pruner~\cite{xie2024moe} prunes weights with the smallest magnitudes, weighted by the corresponding input activations and router weights. The pruned MoE models can benefit from a pre-trained teacher model through expert-wise knowledge distillation and have compatibility with structured pruning. Moreover, STUN~\cite{lee2024stun} combines structured and unstructured pruning to achieve better performance than unstructured pruning alone, utilizing the sparse character of MoEs based on behavior similarity in a greedy manner. MoE-Compression~\cite{he2024demystifying} proposes a unified framework for MoE compression, using both structured and unstructured pruning methods to achieve significant inference speedup with minimal accuracy loss. 

Another potential approach is merging outdated experts based on their parameters to achieve better performance. Branch-Train-Merge~\cite{li2022branch} independently trains different parts of the model on distinct subsets of data, eliminating the need for large-scale multi-node synchronization typically required in traditional LLM training. Building on this, Branch-Train-Mix~\cite{sukhbaatar2024branch} trains multiple copies of a seed LLM to specialize in multiple domains in an asynchronous and parallel fashion, then merges the parameters of the MoE layer to create a unified model that can be further trained. The second finetuning stage makes the final LLM more performant. They also find that their approach is more computing efficient compared to the dense model or specialized MoE model on training. Park et al.~\cite{park2024learning} observed that introducing a shared layer in the MoE could lead to performance degradation. To address this, they trained merged experts that learned the same features in different ways, improving their ability to generalize and mitigating catastrophic forgetting during incremental learning of multi-domain tasks. MC-SMoE~\cite{li2023merge} divides experts into different groups based on routing policies and then merges each group into one expert. HC-SMoE~\cite{chen2024retraining} is a framework that uses hierarchical clustering to merge experts without requiring retraining and can be applied in a task-agnostic manner. During inference, MEO~\cite{he2023merging} systematically investigates the computational cost of MoE. In the drop-in replacement algorithm, they first merge the selected expert parameters and then perform efficient computation. FoE~\cite{wang2023fusing} fuses the outputs of expert models by leveraging their complementary knowledge of the data distribution, framing it as a supervised learning instance. DEK~\cite{zhang2024diversifying} identifies and groups similar experts in feature space, then merges experts within the same group in weight space to reduce the expert count. EEP~\cite{liu2024efficient} introduces a two-stage approach to both prune and merge experts, reducing the total number of experts (saving GPU memory) and the number of active experts (accelerating inference). Inspired by the concept of knowledge transfer in multi-task learning, HyperMoE~\cite{zhao-etal-2024-hypermoe} proposes a novel expert network that further increases sparsity while utilizing information from unselected experts as supplementary input. In practice, they capture the contextual information of experts to compensate for the performance loss of transferring knowledge to specific experts. LiteMoE~\cite{zhuang2024litemoe} retains the most critical experts based on the application's characteristics, merges secondary experts, and obtains the final sparse model without retraining. This approach enables efficient deployment of lightweight submodels on resource-constrained mobile devices.

\subsubsection{Expert Quantization}

\begin{table}[]
\begin{tabular}{c|cc|c|c|c|c}
\hline
\multirow{2}{*}{Method} & \multicolumn{2}{c|}{Quantization Type}                        & \multirow{2}{*}{\begin{tabular}[c]{@{}c@{}}Memory \\ Reduction\end{tabular}} & \multicolumn{1}{c|}{\multirow{2}{*}{\begin{tabular}[c]{@{}c@{}}Accuracy \\ Drop\end{tabular}}} & \multicolumn{1}{c|}{\multirow{2}{*}{\begin{tabular}[c]{@{}c@{}}Inference \\ Speedup\end{tabular}}} & \multicolumn{1}{c}{\multirow{2}{*}{\begin{tabular}[c]{@{}c@{}}Quantization \\ Bits\end{tabular}}}\\ \cline{2-3}
                        & \multicolumn{1}{l|}{Weight} & \multicolumn{1}{l|}{Activation} &                                   &                                &                                    &                                    \\ \hline
MC-MoE~\cite{li2023merge}                  & \multicolumn{1}{c|}{\ding{51} }       &                                 & 4.27x                             & 3.8\%                           & 1.80x                              & 1, 2, 3                            \\
MoE-CSP~\cite{kim2022says}                  & \multicolumn{1}{c|}{\ding{51} }       &                                 & 4.00x                             & -                          & 26.00x                             & 4, 8                               \\
MoQE~\cite{kim2023mixture}                    & \multicolumn{1}{c|}{\ding{51} }       &                                 & 4.90x                             & 0.97\%                          & -                              & 2, 3, 4                                  \\
QMoE~\cite{frantar2023qmoe}                    & \multicolumn{1}{c|}{\ding{51} }       &                                 & 20x                               & 6.7\%                           & 0.95x                              & 1, 2                               \\
CMoE~\cite{yuancompressed}                    & \multicolumn{1}{c|}{\ding{51} }       &    \ding{51}                              & 150x                              & 23.81\%                         &         -                           & 1, 2, 4                            \\
MoE-MPTQS~\cite{imani2024mixture}               & \multicolumn{1}{c|}{\ding{51} }       &                                 & -                                  & 0  $\sim$ 4.98\%                        & 1.00x $\sim$ 20.63x                       & 4, 8                               \\
HOBBIT~\cite{tang2024hobbit}                  & \multicolumn{1}{c|}{\ding{51} }       &                                &     -                              & 1\%                             & 1.35x                              & 2, 4                               \\
EdgeMoE~\cite{yi2023edgemoe}                 & \multicolumn{1}{c|}{\ding{51} }       &                                 & 1.05x  $\sim$ 1.18x                       & 5\%                             & 1.11x  $\sim$ 2.78x                        & 2, 4, 8                            \\ \hline
\end{tabular}
\caption{A comparison of quantization methods.}
\label{tab:expert-quantization}
\end{table}

In addition to model pruning, quantization is an effective technique for reducing model size by converting high-precision weights into low-precision. Given the redundancy of experts in MoE models, current research primarily focuses on quantizing the weights of the experts in MoE. As illustrated in Figure~\ref{fig:compression}-(b), experts are quantized into appropriate low-precision versions using various methods. MC-MoE~\cite{huang2024mc} leverages the access frequency and activation weight to assess the importance of each expert. Along with the associated quantization loss, these metrics are used to determine the optimal quantization configuration for experts via the Integer Programming model. Specifically, the $i$-th expert's access frequency is defined as $\phi_i = \frac{n_i}{N}$, and the activation weight is defined as $w_i = \frac{\sum_{j=1}^N \sigma_j}{N}$, where $n_i$ is the usage frequency, $\sigma_j$ is the routing weight, and $N$ represents the size of the calibration dataset. The quantization loss, $\epsilon_{ij}$ (computed using the Frobenius norm), is then determined for quantizing expert $i$ to $j$ bits ($j \in {1,2,3}$). Using these metrics, MC-MoE defines the objective function as $\sum_{i=1}^n\sum_{j=1}^3 \phi_i^{\alpha} \cdot w_{i}^{\beta} \cdot (\epsilon_{ij}\cdot x_{ij})^{\gamma}$, which is minimized using Integer Programming to determine the optimal bit-width for each expert. MoE-CSP~\cite{kim2022says} quantizes expert weights to either 4 or 8 bits to reduce memory consumption in MoE models. Additionally, it designs specific CUDA kernels that handle the 4-bit/8-bit quantized weights and perform floating-point calculations to accelerate computations. MoQE~\cite{kim2023mixture} quantizes expert weights to 2 bits to address the memory and latency challenges of MoE models based on its observations that quantizing expert FFN layers to 2 bits does not significantly affect model quality, while quantizing other components, like self-attention, significantly hurts performance. Further advancements include QMoE~\cite{frantar2023qmoe} and CMoE~\cite{yuancompressed}, which aggressively compress MoE model into just 1 bit. QMoE implements a highly scalable compression algorithm for large models and introduces a custom compression format along with bespoke GPU kernels for efficient on-the-fly computation. On the other hand, CMoE uses binary-weight networks to quantize model weights to 1 bit and applies learned step-size quantization to activations, quantizing them to 4 bits for MoE-based ASR models, enabling deployment on embedded devices. Moreover, some MoE-optimized systems leverage quantization for better system efficiency. MoE-MPTQS~\cite{imani2024mixture} and HOBBIT~\cite{tang2024hobbit} dynamically select quantized experts to replace original experts based on the current inputs, thereby reducing the expert loading cost on memory-limited devices. EdgeMoE~\cite{yi2023edgemoe} statistically determines the appropriate expert bit-width by profiling expert importance on a calibration dataset. Furthermore, QMoE-Benchmark~\cite{li2024examining} provides a benchmark for exploring various MoE structure-aware quantization heuristics, from coarse to fine granularity. The study reveals that different MoE structures (e.g., blocks, experts, linear layers) require different numbers of weight bits for effective and efficient quantization.

We summarize the main results reported by the methods discussed above in Table~\ref{tab:expert-quantization}. From the table, we observe that quantization primarily benefits memory reduction, with most methods achieving more than a 4x reduction in memory usage. Some methods also lead to actual inference speedup, while others do not. For instance, QMoE even incurs a 5\% overhead due to the absence of a dedicated 1-bit inference CUDA kernel implementation and hardware support. Furthermore, quantization typically causes some accuracy loss, with lower bit widths resulting in greater accuracy degradation. Therefore, when applying quantization to a model, it is crucial to strike a balance between memory consumption, accuracy, and inference speedup, taking into account the specific requirements and available hardware resources.

\subsubsection{Expert Distillation}

Knowledge distillation is another effective method for creating smaller, yet powerful models from larger ones. As shown in Figure~\ref{fig:compression}-(c), knowledge distillation presents a promising solution to generate a compact, high-performance model from the original MoE model.
LLaVA-MoD~\cite{shu2024llava} combines the MoE structure with knowledge distillation to efficiently train small multimodal large language models (s-MLLMs) from large ones (l-MLLMs). It first incorporates the MoE structure into the s-MLLM to balance computational efficiency and model performance. Then, it introduces two consecutive distillation stages, mimic distillation and preference distillation, to train the s-MLLM. Mimic distillation minimizes the Kullback-Leibler (KL) divergence between the output distributions of the s-MLLM and l-MLLM, enabling the s-MLLM to emulate the l-MLLM’s understanding. Preference distillation further refines the s-MLLM using Preference Optimization with additional datasets.
DeepSpeed-MoE~\cite{DeepSpeedmoe} employs staged knowledge distillation to create a distilled version of its proposed PR-MoE model, called MoS. This method reduces model size while maintaining performance. Additionally, some studies focus on transferring the power of sparse models to dense models through knowledge distillation for more efficient deployment. For example, OneS~\cite{xue2022one} generates a dense model from a MoE model in two steps: knowledge gathering and knowledge distillation. Knowledge gathering merges experts into a single expert using four aggregation methods, including summation, averaging, top-k knowledge gathering, and Singular Value Decomposition (SVD) knowledge gathering. In the second step, knowledge distillation distills the merged model using the original MoE model. 
What's more, MoE-KD~\cite{salinas2022knowledge} and CMoE~\cite{yuancompressed} distill MoE-based speech recognition models into dense models, accelerating the speech recognition process. Specifically, MoE-KD initializes the FFN module of the student dense model with the most frequently used expert from the teacher MoE model, then trains the FFN modules of the student model through layer-wise distillation. CMoE distills a binary dense model from the original model using quantization techniques. Switch Transformers~\cite{fedus2022switch} and ELSM~\cite{artetxe2021efficient} also explore distillation techniques to convert their sparse models into dense models.
Additionally, to simplify the construction of MoE models, LaDiMo~\cite{kim2024ladimo} converts a dense model into a sparse MoE model via layer-wise distillation, where each expert learns to approximate the results of the original dense layers.

\subsubsection{Expert Decomposition}

As shown in Figure~\ref{fig:compression}-(d), low-rank decomposition is an effective method for reducing model size by decomposing a large weight matrix into smaller matrices. MPOE~\cite{gao2022parameter} employs the matrix product operator (MPO), a tensor decomposition technique derived from quantum many-body physics, to decompose the expert weight matrix into a central tensor and several auxiliary tensors. The central tensor retains most of the parameters and core information of the original weight matrix, while the auxiliary tensors are much smaller and serve as complements to the central tensor. After decomposition, all experts in a layer share the same central tensor, thereby significantly reducing the total number of parameters in that layer.
MC-SMoE~\cite{li2023merge} first groups the experts into several clusters and then merges each group into a single expert using a weighted sum. Low-rank decomposition is then applied to the merged experts to further reduce the model size. This approach is based on the observation that the merged experts have a lower rank compared to the original experts, making them more suitable for decomposition.
MoE-$I^{2}$~\cite{yang2024moe} identifies the importance of each expert and assigns higher ranks to more important experts while assigning lower ranks to less important ones for low-rank decomposition. The importance, \(I_{i,j}\), of the \(j\)-th expert in the \(i\)-th layer, \(e_{i,j}\), is determined by removing \(e_{i,j}\) and calculating the performance loss compared to the original model. The SVD rank, \(r_{i,j}\), for \(e_{i,j}\) is then computed as $r_{i,j} = \left\lfloor \frac{(I_{i,j} + \epsilon)^{\alpha}}{\sum_{j=1}^{M_i} (I_{i,j} + \epsilon)^{\alpha}} \right\rfloor \cdot R_a \cdot M_i$, where \(\epsilon\) and \(\alpha\) are hyperparameters, \(R_a\) is the compression ratio, and \(M_i\) is the number of experts in layer \(i\).

\subsection{Algorithm Improvement}
In this part, we conduct an in-depth exploration of two other strategies for improving the MoE model inference algorithm.

\begin{figure}[t]
    \centering
    \subfloat[Experts Skipping]{\includegraphics[width=0.41\linewidth]{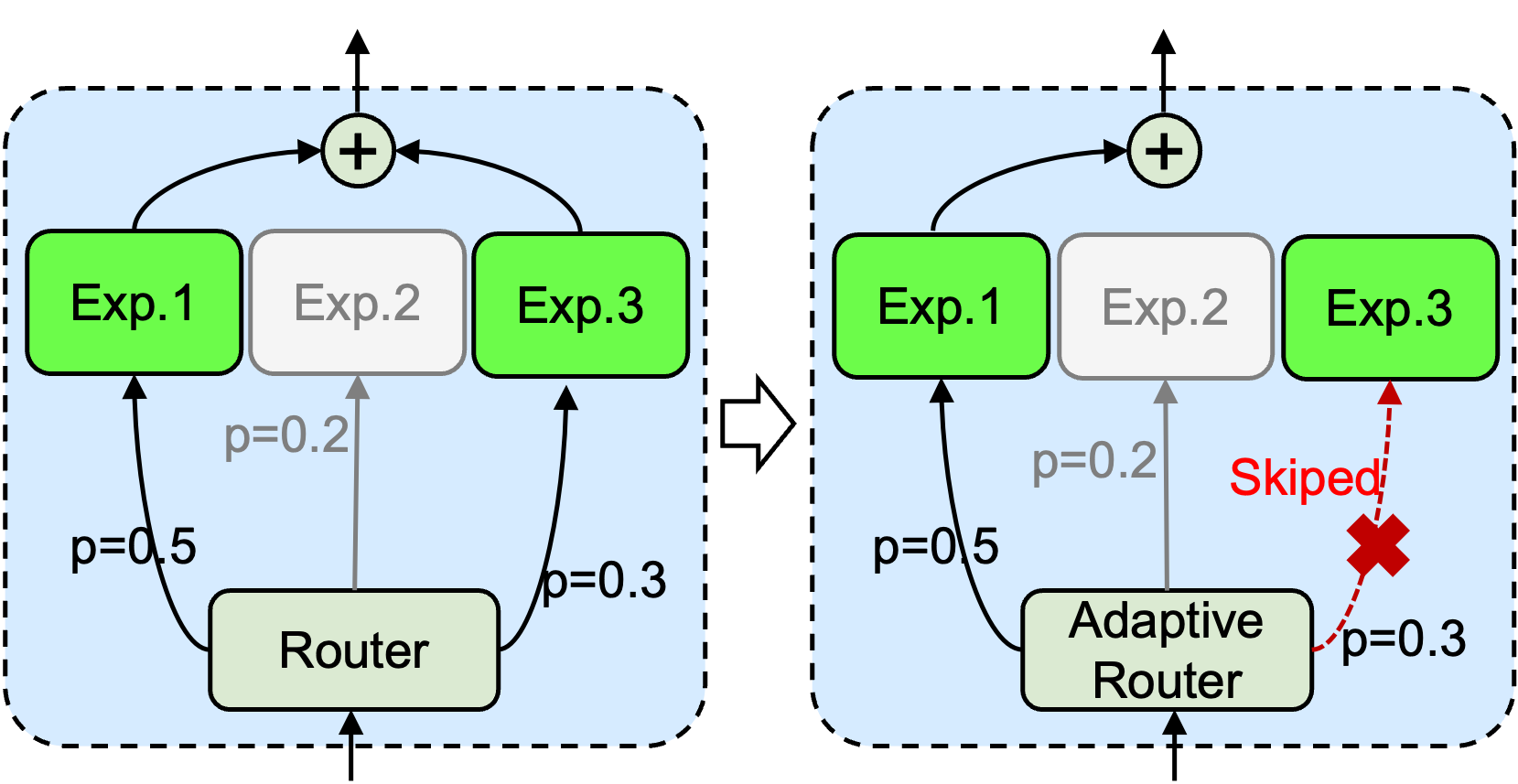}}
    \hspace{30pt}
    \subfloat[Sparse to Dense]{\includegraphics[width=0.29\linewidth]{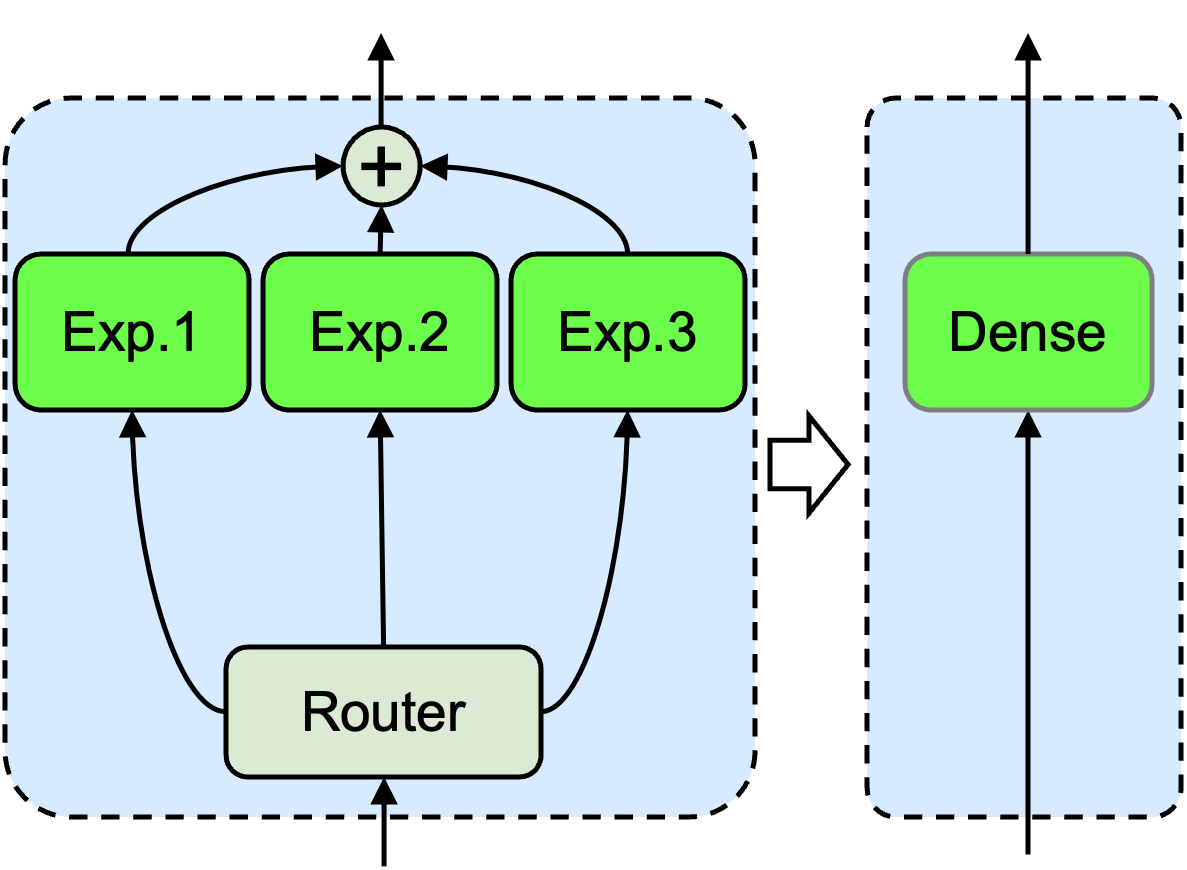}}
    
    \caption{Algorithm improvement on expert layers.}
  \label{fig:algorithmtechs}
\end{figure}

\subsubsection{Dynamic Gating}

Given the significant progress made in utilizing the sparsity of MoE models, specific strategies can be employed to further exploit this sparsity in the inference process. Due to the vertical parallel structure of MoE experts, dynamically reducing the number of experts activated for each token clearly presents an effective strategy. Figure~\ref{fig:algorithmtechs}-(a) illustrates the MoE layer's calculation process after the skip mechanism is applied, in contrast to the original process.

Li \textit{et al.}~\cite{li2023adaptive} proposes a self-adaptive gating mechanism that dynamically determines the number of experts required for each token, based on the expert probability distribution. This method enhances training efficiency while preserving the sparsity of the MoE model and further reduces training time through curriculum learning. DynMoE~\cite{guo2024dynamic} introduces two innovative methods for expert activation: a top-down gating approach that enables flexible per-token expert allocation, and an adaptive mechanism that dynamically determines the number of experts needed for each token based on computational requirements.
XMoE~\cite{yang2024xmoe} employs a strategy of using smaller, but more experts, with dynamic expert activation based on threshold values to balance computational efficiency and model performance. AdapMoE~\cite{zhong2024adapmoe} is a co-design framework aimed at improving inference efficiency on edge devices. It adaptively applies three techniques: expert activation, expert prefetching, and GPU cache allocation. The comprehensive comparison of performance and methods of these works is presented in Table~\ref{tab:dynamic gating}. DA-MoE~\cite{aghdam2024damoedynamicexpertallocation}  introduces a token importance prediction method derived from attention mechanisms to guide expert allocation. By assigning experts based on token importance scores, their approach achieves strong performance on the GLUE benchmark while demonstrating effective scaling with increased expert count.

\begin{table}[]
\centering 
\begin{tabular}{llllll}
\hline
\textbf{Method}                                                                          & \textbf{FLOPs Reduction} & \textbf{Speedup} & \textbf{Threshold Strategy}         & \textbf{Load Balance}               & \textbf{PR.}      \\ \hline
Fixed top-k gating                                                              & 0\%             & 1.0x    & \ding{55} & \ding{55} & \ding{51} \\
Li et al.\cite{li2023adaptive} & 38.2\%          & 1.32x   & accumulative probability   & soft on top-1              & \ding{51} \\
DynMoE\cite{guo2024dynamic}                              & 9\%             & 1.37x   & single expert probability  & \ding{51} & \ding{55} \\
XMoE\cite{yang2024xmoe}                                  & 75\%            & -       & accumulative probability   & \ding{51} & \ding{51} \\
AdapMoE\cite{zhong2024adapmoe}                           & 25\% of experts & 1.35x   & performance perturbation   & \ding{55} & \ding{51} \\ \hline
\end{tabular}
\caption{A comparison of dynamic gating methods. PR. indicates performance Retention. Specially, Li et al.~\cite{li2023adaptive} only uses soft load balance constraints on top-1 gating. No specific acceleration ratio was provided in XMoE~\cite{yang2024xmoe}.}\label{tab:dynamic gating}
\end{table}

\subsubsection{Sparse to Dense}

In certain scenarios, dense models offer unique advantages due to their smaller number of parameters. Therefore, transforming MoE models into dense target models all at once can achieve maximum sparsity while maintaining model performance. Most approaches use knowledge distillation to achieve this sparse-to-dense conversion, as illustrated in Figure~\ref{fig:algorithmtechs}-(b).

XFT~\cite{ding2024xft} proposes a novel method to supervise fine-tune dense LLMs. They first generate a sparse-upcycled MoE model, and then transform it back into an efficient dense LLM of the same size and structure through a learnable merging mechanism. Switch Transformers~\cite{fedus2022switch} explores distillation techniques to convert a sparse model into a dense one, demonstrating that the dense model can retain over 30\% of its performance even after compressing 97\% of the parameters from the sparse MoE model. 
ELSM (Artetxe et al., 2021) demonstrates that dense student models distilled from sparse MoE teachers can match and even surpass the teacher's performance. OneS~\cite{xue2022one} employs four distinct methods to generate the dense model, including summation, averaging, top-k Knowledge Gathering, and Singular Value Decomposition Knowledge Gathering. The model is then refined to reduce noise by gathering sparse knowledge. TSEP~\cite{chen2022task} progressively eliminates non-expert components based on specific downstream tasks, ultimately converting the sparse MoE model into a dense counterpart. EWA~\cite{huang2023experts} replaces FFNs with specially designed MoEs during training before reverting to dense ViT for inference. AdaMoLE~\cite{liu2024adamolefinetuninglargelanguage} combining the Low-Rank Adaptation (LoRA) structure, a dedicated network is used to adjust the activation threshold for different task complexities. It has shown superior performance to baseline in many natural language processing tasks, especially in some commonsense reasoning tasks.

\section{System-level Optimizations}\label{sec:system}

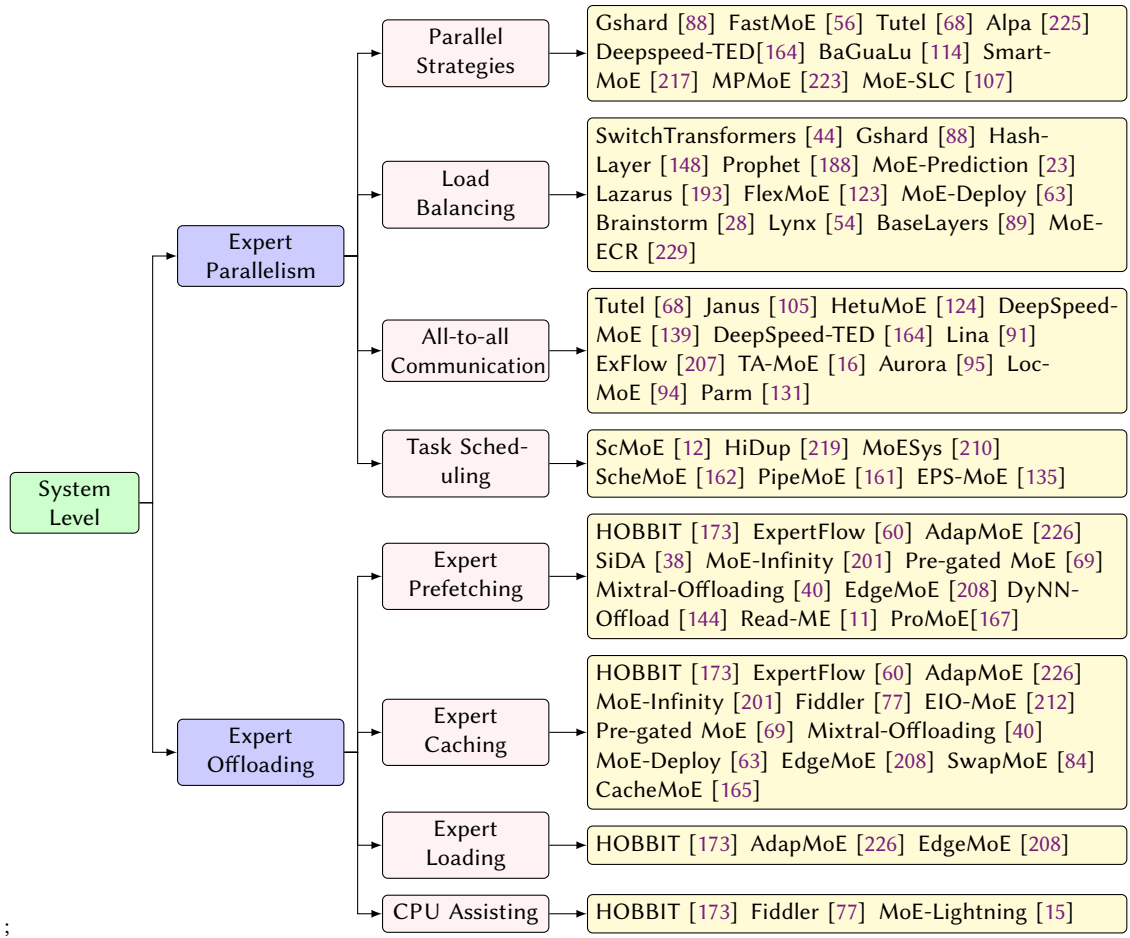
\begin{figure}
    \centering
    
\tikzset{
    basic/.style  = {draw, text width=3cm, align=center, font=\sffamily, rectangle},
    root/.style   = {basic, rounded corners=2pt, thin, align=center, text width=1.5cm, fill=red!20},
    onode/.style = {basic, thin, rounded corners=2pt, align=center, fill=green!20,text width=1.5cm,},
    xnode/.style = {basic, thin, rounded corners=2pt, align=center, fill=blue!20,text width=2cm,},
    wnode/.style = {basic, thin, rounded corners=2pt, align=center, fill=pink!20, text width=2cm},
    tnode/.style = {basic, thin, rounded corners=2pt, align=left, fill=yellow!20, text width=29em},
    ttnode/.style = {basic, thin, rounded corners=2pt, align=left, fill=yellow!20, text width=22em},
    edge from parent/.style={draw=black, edge from parent fork right}

}

\begin{forest} for tree={
    grow'=east,
    growth parent anchor=west,
    parent anchor=east,
    child anchor=west,
    edge path={\noexpand\path[\forestoption{edge},->, >={latex}] 
         (!u.parent anchor) -- +(5pt,0pt) |-  (.child anchor) 
         \forestoption{edge label};},
}
[System Level, onode,  l sep=5mm, before typesetting nodes={if n=1{anchor=south}{ };},
    [Expert Parallelism, xnode, l sep=5mm,
        [Parallel Strategies, wnode, l sep=5mm,calign=child edge,
            [Gshard~\cite{lepikhin2020gshard}\, FastMoE~\cite{he2021fastmoe}\, Tutel~\cite{hwang2023tutel}\, Alpa~\cite{Alpa}\, Deepspeed-TED\cite{DeepSpeedted}\, BaGuaLu~\cite{BaGuaLu}\, SmartMoE~\cite{288691}\, MPMoE~\cite{10494556}\, MoE-SLC~\cite{liu2025optimizing}, ttnode]
        ]
        [Load \\Balancing, wnode, l sep=5mm,calign=child edge,
            [SwitchTransformers~\cite{fedus2022switch}\, Gshard~\cite{lepikhin2020gshard}\, HashLayer~\cite{roller2021hash}\, Prophet~\cite{10319949}\, MoE-Prediction~\cite{cong2024prediction}\, Lazarus~\cite{wu2024lazarus}\, FlexMoE~\cite{nie2023flexmoe}\, MoE-Deploy~\cite{huang2023towards}\, Brainstorm~\cite{Brainstorm}\, Lynx~\cite{gupta2024lynx}\, BaseLayers~\cite{lewis2021base}\, MoE-ECR~\cite{moeecr},  ttnode]
        ]
        [All-to-all\\ Communication, wnode, l sep=5mm,calign=child edge,
            [Tutel~\cite{hwang2023tutel}\, Janus~\cite{10.1145/3603269.3604869}\, HetuMoE~\cite{nie2022hetumoe}\, DeepSpeed-MoE~\cite{DeepSpeedmoe}\, DeepSpeed-TED~\cite{DeepSpeedted}\, Lina~\cite{288705}\, ExFlow~\cite{yao2024exploiting}\, TA-MoE~\cite{chen2022ta}\, Aurora~\cite{li2024optimizing}\, LocMoE~\cite{li2024locmoe}\, Parm~\cite{10621327}, ttnode]
        ]
        [Task Scheduling, wnode, l sep=5mm,calign=child edge,
            [ScMoE~\cite{cai2024shortcut}\, HiDup~\cite{HiDup}\, MoESys~\cite{10528887}\, ScheMoE~\cite{10.1145/3627703.3650083}\, PipeMoE~\cite{10228874}\, EPS-MoE~\cite{qian2024eps}, ttnode]
        ]
    ]
    [Expert Offloading, xnode, l sep=5mm,
        [Expert Prefetching, wnode, l sep=5mm,calign=child edge,
            [HOBBIT~\cite{tang2024hobbit}\, ExpertFlow~\cite{he2024expertflow}\, AdapMoE~\cite{zhong2024adapmoe}\, SiDA~\cite{du2024sida}\, MoE-Infinity~\cite{xue2024moe}\, Pre-gated MoE~\cite{hwang2024pre}\, Mixtral-Offloading~\cite{eliseev2023fast}\, EdgeMoE~\cite{yi2023edgemoe} DyNN-Offload~\cite{dynnoffload}\, Read-ME~\cite{readmoe}\, ProMoE\cite{song2024promoe},  ttnode]
        ]
        [Expert Caching, wnode, l sep=5mm,calign=child edge,
            [HOBBIT~\cite{tang2024hobbit}\, ExpertFlow~\cite{he2024expertflow}\, AdapMoE~\cite{zhong2024adapmoe}\, MoE-Infinity~\cite{xue2024moe}\, Fiddler~\cite{kamahori2024fiddler}\, EIO-MoE~\cite{yuan2024efficient}\, Pre-gated MoE~\cite{hwang2024pre}\, Mixtral-Offloading~\cite{eliseev2023fast}\, MoE-Deploy~\cite{huang2023towards}\, EdgeMoE~\cite{yi2023edgemoe}\, SwapMoE~\cite{kong2023serving}\, CacheMoE~\cite{skliar2024mixture},  ttnode]
        ]
        [Expert Loading, wnode, l sep=5mm,calign=child edge,
            [HOBBIT~\cite{tang2024hobbit}\, AdapMoE~\cite{zhong2024adapmoe}\, EdgeMoE~\cite{yi2023edgemoe}, ttnode]
        ]
        [CPU Assisting, wnode, l sep=5mm,calign=child edge,
            [HOBBIT~\cite{tang2024hobbit}\, Fiddler~\cite{kamahori2024fiddler}\,  MoE-Lightning~\cite{cao2024moelight}, ttnode]
        ]
    ] 
]
\end{forest}
    \caption{System-level inference optimization techniques for MoE.}
    \label{fig:systeml-opt}
\end{figure}

Due to the unique structure of MoE, many studies focus on accelerating inference at the system level by leveraging the sparse activation patterns inherent to this architecture. Typically, MoE models are deployed in two scenarios: cloud environments with multiple powerful servers and edge environments with a single device. 
In cloud clusters, MoE models are distributed across many devices to enable parallel execution. In addition to traditional parallelization techniques such as data parallelism, tensor parallelism, and pipeline parallelism~\cite{narayanan2021efficient, rajbhandari2020zero, jia2019beyond}, expert parallelism is a specialized method tailored specifically for MoE models. 
On edge devices, limited GPU memory often cannot accommodate all parameters of an MoE model, necessitating the offloading of some parameters to CPU memory or SSD storage. To address this, the expert-offloading technique has been developed to fully utilize the sparse activation patterns of experts for efficient execution.  Figure~\ref{fig:systeml-opt} shows the detailed structure of this
section.

\subsection{Expert Parallelism}

\begin{figure}[t]
    \centering
    \subfloat[Expert Parallelism]{\includegraphics[width=0.56\linewidth]{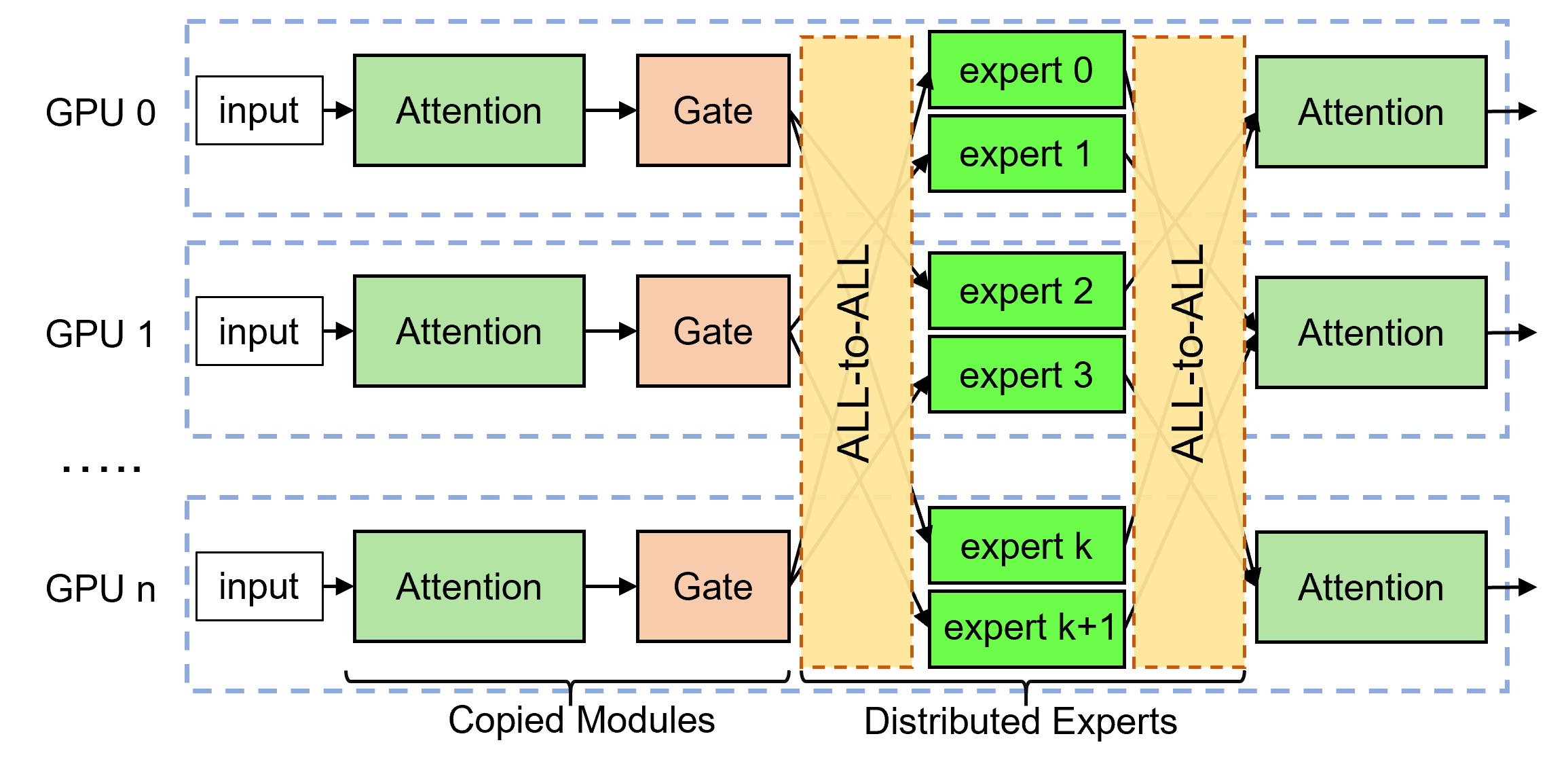}}
    \subfloat[Expert Offloading]{\includegraphics[width=0.44\linewidth]{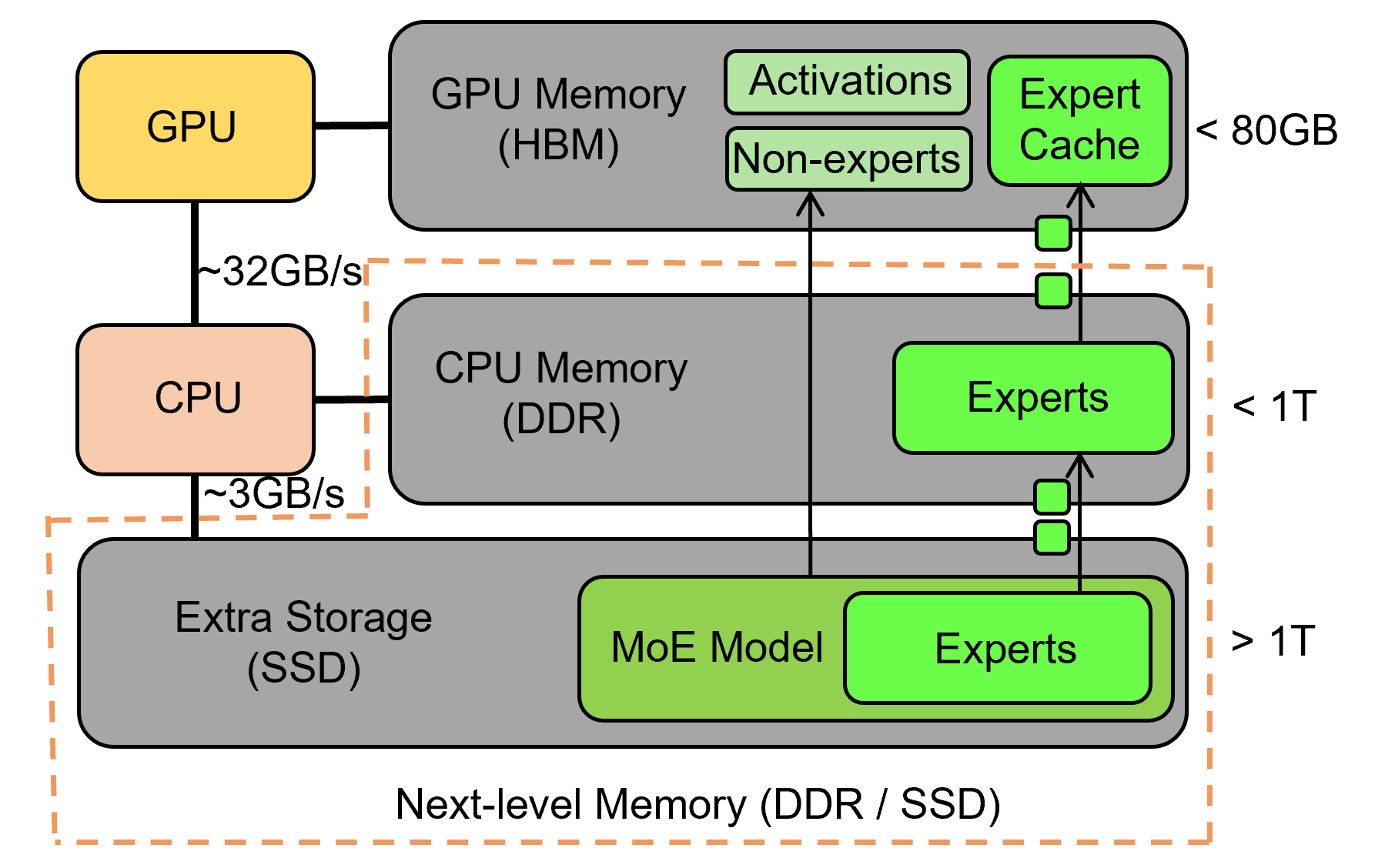}}
    \caption{Expert parallelism and expert offloading techniques.}
  \label{fig:system-level-op}
\end{figure}

Expert parallelism is a key technique for deploying large MoE models across multiple devices to enable distributed execution. As Figure~\ref{fig:system-level-op}-(a) shown, when distributing MoE layers, each device holds a subset of experts along with all other non-expert parameters~\cite{10.1145/3627703.3650083}. In special cases where an expert is very large, each device may contain only one expert. During the execution of an MoE layer, inputs are processed on each device by the attention module and gate module. Subsequently, an all-to-all communication operation redistributes tokens to corresponding devices based on the gate output, allowing each expert to process its assigned inputs. Finally, another all-to-all communication operation occurs to exchange the experts’ outputs and send them back to the originating devices. Consequently, the execution time of MoE layers is dominated by two primary phases: computation and communication. To accelerate inference, various studies have proposed optimizations for these phases.

Existing research focuses on optimizing MoE training and inference performance along four main dimensions: parallelism strategy, load balancing, all-to-all communication, and task scheduling.

\subsubsection{Parallelism Strategy Design}

In addition to relying solely on expert parallelism~\cite{he2021fastmoe, lepikhin2020gshard}, recent studies combine multiple parallel strategies to maximize parallelism and achieve more efficient distributed computing.
Tutel~\cite{hwang2023tutel} dynamically switches parallelism strategies at each iteration without incurring extra overhead. Specifically, this work employs a single distribution layout that encompasses all optimal strategies, thereby eliminating the need to reformat input data or weights when switching parallelism strategies.
Alpa~\cite{Alpa} reclassifies conventional data and model parallelism into intra-operator and inter-operator parallelism based on the key observation that different parallelization techniques have varying bandwidth requirements for communication. In typical clusters, co-located devices have high communication bandwidth, whereas distant devices have limited communication bandwidth. With this reclassification, Alpa can automatically derive efficient parallel execution plans that incorporate data, operator, and pipeline parallelism based on the model description and cluster configuration.
DeepSpeed-TED~\cite{DeepSpeedted} implements a hybrid parallel algorithm combining data, tensor, and expert parallelism. It designs a tiled version of the optimizer to reduce peak memory consumption arising from the combination of these three parallelism forms, thereby enabling the training of MoE models with 4–8× larger base models than the current state-of-the-art.
BaGuaLu~\cite{BaGuaLu} proposes a parallel strategy called MoDa that combines MoE parallelism and data parallelism to effectively scale models. Specifically, MoDa places data parallelism within a single supernode while maintaining MoE communication globally across supernodes. This approach is based on the observation that All-reduce communication (in data parallelism) incurs much higher costs than All-to-all communication (in MoE parallelism) during the training of its target large-scale models.
SmartMoE~\cite{288691} explores the space of hybrid parallelism with an awareness of heterogeneous workloads, incorporating potentially faster candidate parallel strategies. It introduces a two-stage approach to identify optimal parallel strategies for data-sensitive models, periodically determining whether a new parallel strategy should be employed at runtime.
MPMoE~\cite{10494556} proposes a lightweight profile-based algorithm that leverages profiling information to identify the most suitable configuration at runtime, optimizing pipeline parallelism. Additionally, in serverless  computing,  MoE-SLC~\cite{liu2025optimizing} proposes a Bayesian optimization framework with multi-dimensional $\delta$-greedy search to learn expert selections for expert parallelism and optimize the MoE deployment to achieve optimal billed cost.

\subsubsection{Load Balancing}

Token processing in MoE models can lead to imbalances, where some experts receive significantly more tokens than others. This disparity results in certain devices experiencing high overheads while others remain idle, ultimately delaying overall computation.
To address this issue, some works propose adding auxiliary load balancing loss terms to train the gate module~\cite{shazeer2017outrageously, fedus2022switch, lepikhin2020gshard}, while others implement gate modules with special hash functions to prevent load imbalance~\cite{roller2021hash}. Advanced techniques focus on optimizing expert placement.
For example, Prophet\cite{10319949} builds a load balance performance model to estimate the execution time of the MoE layer for each expert placement and then uses a greedy search algorithm to find a well-balanced expert placement. MoE-Prediction\cite{cong2024prediction} traces and analyzes the loads of each expert during the training process, then employs classical prediction algorithms to forecast the expert load proportions, providing valuable guidance for expert placement in MoE training. Lazarus\cite{wu2024lazarus} allocates expert replicas using an optimal placement algorithm that assigns more replicas to popular experts to help balance the load. FlexMoE\cite{nie2023flexmoe} employs fine-grained replication strategies that select specific heavy experts and duplicate them across multiple devices, dynamically adjusting the expert-to-device mapping at runtime to address routing imbalances.
Additional strategies, such as those proposed by MoE-Deploy\cite{huang2023towards} and Brainstorm\cite{Brainstorm}, rely on historical allocation data and expert reordering to balance loads. Furthermore, techniques like Lynx\cite{gupta2024lynx} reduce the number of active experts during batch inference, and BaseLayers\cite{lewis2021base} formulates token-to-expert allocation as a linear assignment problem to ensure equitable token distribution. Additionally, MoE-ECR~\cite{moeecr} allows experts to select the top-k tokens instead of letting each token choose the top-k experts, ensuring that each expert has a fixed bucket size.

\subsubsection{All-to-all Communication Optimization}

All-to-all communication is a significant bottleneck in MoE layer execution~\cite{li2024optimizing, 10.1145/3503221.3508418, DeepSpeedmoe}, prompting substantial research aimed at optimizing this operation.
Many efforts focus on hierarchical communication strategies and data compression techniques to reduce overhead. For instance, Tutel\cite{hwang2023tutel}, HetuMoE\cite{nie2022hetumoe}, and DeepSpeed-MoE~\cite{DeepSpeedmoe} utilize the hierarchical all-to-all algorithm to optimize communication by combining intra-node and inter-node hierarchies. Compared to traditional all-to-all operations, which allow all GPUs to communicate directly with one another, this hierarchical approach first gathers data from all GPUs within a node to a single GPU (intra-node) and then communicates between nodes (inter-node). This method speeds up communication by fully utilizing both levels of bandwidth.
Data compression strategies, such as those employed by DeepSpeed-TED\cite{DeepSpeedted} and TA-MoE\cite{chen2022ta}, minimize data transfer by eliminating unnecessary information and adapting data volume to the network topology. Innovative paradigms like Janus~\cite{10.1145/3603269.3604869} adopt a data-centric approach, moving experts between devices instead of tokens, thereby significantly reducing communication costs.
ExFlow\cite{yao2024exploiting} reduces the number of all-to-all operations from two to one by ensuring token context coherence, meaning that all GPUs have the context of all requests rather than only their own. It then exploits inter-layer expert affinity to further reduce the number of all-to-all operations by optimally placing experts in each layer, increasing the probability that tokens remain on local GPUs. Aurora\cite{li2024optimizing} orders token transmission to avoid bandwidth contention, achieving minimal all-to-all communication costs through theoretical derivation.
LocMoE\cite{li2024locmoe} and Parm\cite{10621327} further enhance communication efficiency by converting inter-node communication to intra-node operations and overlapping intra-node with inter-node communications. Additionally, Lina~\cite{288705} accelerates all-to-all communication by prioritizing all-to-all operations over all-reduce operations.

\subsubsection{Task Scheduling}
To overlap communication and computation tasks and reduce end-to-end runtime, various task scheduling strategies have been developed. Advanced scheduling strategies leverage architectural and algorithmic innovations to achieve this overlap effectively.
For example, ScMoE~\cite{cai2024shortcut} introduces a shortcut-connected MoE architecture that processes representations from both the current and preceding layers, rather than solely processing representations from the current layer. Specifically, ScMoE employs a top-1 MoE module to handle representations from the preceding layer via a shortcut connection, while a shared expert processes the current layer's representations. This structure effectively decouples communication from its conventional sequence, thereby enhancing the overlap between communication and computation.
HiDup~\cite{HiDup} splits the input data on each device into two microbatches with no dependencies between them. Consequently, the training of the two microbatches on each device can be carried out in parallel, effectively overlapping the computation of one microbatch with the communication of the other.
MoESys~\cite{10528887} employs an elastic MoE training strategy with 2D prefetching and fusion communication over hierarchical storage to overlap computation with the parameter readiness from the hierarchical storage.
ScheMoE\cite{10.1145/3627703.3650083} first modularizes the computing tasks (data compression and expert computation) and communication tasks (collective communication). It then designs an adaptive optimal scheduling algorithm to pipeline the communication and computing tasks based on these modularized operations. 
Similarly, PipeMoE\cite{10228874} designs a performance model to predict the computation and communication costs for various MoE workloads and then proposes an optimal polynomial-time solution to pipeline tasks, thereby hiding communication latency based on the performance model's results.
Additionally, EPS-MoE~\cite{qian2024eps} dynamically selects optimal kernel implementations to adaptively overlap FFN module computation with all-to-all communication, further improving runtime efficiency.

\begin{table}[]
\begin{tabular}{ccc|ccc}
\toprule
\multicolumn{3}{c|}{DeepSpeed-MoE~\cite{DeepSpeedmoe}}                                                     & \multicolumn{3}{c}{FasterMoE~\cite{10.1145/3503221.3508418}}                                                    \\ \midrule
\multicolumn{1}{c}{Method}        & \multicolumn{1}{c}{Speedup} & Metric             & \multicolumn{1}{c}{Method}   & \multicolumn{1}{c}{Speedup} & Metric             \\ \midrule
\multicolumn{1}{c}{HetuMoE~\cite{nie2022hetumoe}}       & \multicolumn{1}{c}{5.88x}   & Time per iteration & \multicolumn{1}{c}{TA-MoE~\cite{chen2022ta}}   & \multicolumn{1}{c}{1.39x}   & Tokens per second   \\ 
\multicolumn{1}{c}{TA-MoE~\cite{chen2022ta}}        & \multicolumn{1}{c}{1.31x}   & Tokens per second   & \multicolumn{1}{c}{PipeMoE~\cite{10228874}}  & \multicolumn{1}{c}{1.69x}   & Time per iteration \\ 
\multicolumn{1}{c}{FlexMoE~\cite{nie2023flexmoe}}       & \multicolumn{1}{c}{1.70x}   & Time to converge   & \multicolumn{1}{c}{FlexMoE~\cite{nie2023flexmoe}}  & \multicolumn{1}{c}{1.30x}   & Time to converge  \\  
\multicolumn{1}{c}{Prophet~\cite{10319949}}       & \multicolumn{1}{c}{2.39x}   & Time per iteration & \multicolumn{1}{c}{Prophet~\cite{10319949}}  & \multicolumn{1}{c}{1.43x}   & Time per iteration \\ 
\multicolumn{1}{c}{Parm~\cite{10621327}}          & \multicolumn{1}{c}{3.00x}   & Time per iteration & \multicolumn{1}{c}{ScheMoE~\cite{10.1145/3627703.3650083}}  & \multicolumn{1}{c}{1.25x}   & Time per iteration \\ 
\multicolumn{1}{c}{Lazarus~\cite{wu2024lazarus}}       & \multicolumn{1}{c}{3.40x}   & Tokens per second   & \multicolumn{1}{c}{MPMoE~\cite{10494556}}    & \multicolumn{1}{c}{1.60x}   & Time per iteration \\ 
\multicolumn{1}{c}{DeepSpeed-TED~\cite{DeepSpeedted}} & \multicolumn{1}{c}{1.35x}   & Time per iteration & \multicolumn{1}{c}{SMARTMOE~\cite{288691}} & \multicolumn{1}{c}{1.88x}   & Time per iteration  \\
\bottomrule
\end{tabular}
\caption{Speedup of parallelism systems when compared to DeepSpeed-MoE and FasterMoE in MoE training.}
\label{tab:training-speedup}
\end{table}

\begin{table}[]
\resizebox{\textwidth}{!}{
\begin{tabular}{ccccc}
\toprule
Method        & Speedup                    & Metric & Baseline      & Optimized Techniques                          \\ \midrule
Lina~\cite{288705}          & 1.63x                      & Time per batch              & DeepSpeed~\cite{deepspeed}     &  Prioritize all2all over allreduce      \\ 
TUTEL~\cite{hwang2023tutel}         & 2.03x                      & Time per batch              & Fairseq~\cite{fairseq}       & Design an identical layout for weights \\ 
Aurora~\cite{li2024optimizing}        & 1.81x& Time per batch              & Lina~\cite{288705}          & Strategically order token transmissions                         \\ 
ExFlow~\cite{yao2024exploiting}        & 2.2x & Tokens per second            & Deepspeed-MoE~\cite{DeepSpeedmoe} & Reduce two all2all operations into one                        \\ 
Brainstorm~\cite{Brainstorm}    & 3.33x & Time per batch              & Tutel~\cite{hwang2023tutel}         & Preload weight with profiling statics                                \\ 
MoE-Deploy~\cite{huang2023towards}    & 3.32x                      & Tokens per second            & Tutel~\cite{hwang2023tutel}         &  Combine high-used experts with lows                            \\ 
DeepSpeed-MoE~\cite{DeepSpeedmoe} & 7.25x                      & Time per batch              & PyTorch~\cite{pytorch}       & Groupe  tokens with same data path   \\ 
\bottomrule
\end{tabular}
}
\caption{Speedup of parallelism systems in MoE Inference.}
\label{tab:inference-speedup}
\end{table}

\begin{table}[]
\begin{tabular}{ccc}
\toprule
Method        & Benefits                                                                   & Baseline \\ \midrule
Lina~\cite{288705}           & Gain 2.3x reduction in all2all communication                                   & DeepSpeed~\cite{deepspeed}                     \\ 
Prophet~\cite{10319949}       & Improve the load balance by 1.75x to 12.06x                                   & FasterMoE~\cite{10.1145/3503221.3508418}                     \\ 
ScheMoE~\cite{10.1145/3627703.3650083}       & Gain 1.4x to 2x speedup in all2all communication              & NCCLA2A~\cite{nie2022hetumoe}, 2DH-A2A~\cite{hwang2023tutel}              \\ 
MPMoE~\cite{10494556}         &Achieve a 53\% decrease in memory usage                                               & FasterMoE~\cite{10.1145/3503221.3508418}                     \\ 
MoESys~\cite{10528887}        &      Achieve a 18\%   decrease in memory usage            & DeepSpeed~\cite{deepspeed}                     \\ 
Aurora~\cite{li2024optimizing}        & Achieve a 1.57x to 1.72x increase in GPU utilization & Lina~\cite{288705}      \\ 
MoE-Deploy~\cite{huang2023towards}    & Achieve a 79.6\% decrease in memory usage                                             & Tutel~\cite{hwang2023tutel}                         \\ 
DeepSpeed-TED~\cite{DeepSpeedted} & Support 1.09x to 4.8x larger MoE models                                     & DeepSpeed-MoE~\cite{DeepSpeedmoe}                 \\ 
\bottomrule
\end{tabular}
\caption{Additional benefits of existing systems.}
\label{tab:otherbenefits}
\end{table}

\subsubsection{Performance Summarization}

To assess the performance of the methods discussed above, we extract relevant data from their respective papers. Table~\ref{tab:training-speedup} shows the training speedup of various systems relative to DeepSpeed-MoE and FasterMoE, both of which are widely recognized by researchers and serve as excellent starting points for related studies. Although different systems evaluate their methods using various configurations, a rough comparison can still be made due to the shared baselines. For instance, HetuMoE demonstrates exceptional performance, with the highest speedup compared to other systems.
In addition to training speedup, we also report inference speedup and highlight the main optimization techniques of several systems in Table~\ref{tab:inference-speedup}. However, comparing performance across systems is challenging due to differences in their baselines. Beyond speedup, some systems offer additional benefits, such as reduced memory usage. As shown in Table~\ref{tab:otherbenefits}, many systems focus on optimizing memory usage, which is critical given the large size of MoE models. Additionally, some studies also optimize GPU utilization and communication efficiency.

\subsection{Expert Offloading}

When deploying MoE models on edge devices, parameter-offloading techniques offer a potential solution to mitigate the challenge of insufficient GPU memory for storing all model parameters. However, traditional parameter-offloading methods load model parameters layer by layer from CPU memory or SSD, neglecting the sparse activation characteristics of MoE models. This oversight incurs significant overhead in parameter loading, leading to suboptimal inference performance. To address this limitation, many studies have proposed expert-offloading techniques, which take advantage of MoE sparse activation patterns. Instead of loading all experts for a given layer, expert-offloading selectively loads only the required experts, thereby significantly reducing loading latency.

As Figure~\ref{fig:system-level-op}-(b) shown\cite{tang2024hobbit}, expert-offloading operates by storing all non-expert weights and a subset of critical experts in GPU memory (referred to as the "expert cache") while offloading less frequently used experts to CPU memory or SSD (referred to as "next-level memory"). When required experts are missing from the expert cache, they are fetched from next-level memory and loaded into the cache, potentially replacing some existing experts. Various studies have optimized this process using different strategies to achieve superior inference performance.

\subsubsection{Expert Prefetching}

To minimize the waiting time for the required experts, some studies focus on optimizing prefetching techniques that overlap expert loading with GPU computation. Many works predict the experts needed for subsequent layers. For example, Mixtral-Offloading~\cite{eliseev2023fast}, AdapMoE~\cite{zhong2024adapmoe}, and HOBBIT~\cite{tang2024hobbit} use current gating inputs to feed the gating module for the following layers, predicting the required experts and prefetching them in advance. This prediction is based on the key observation that the inputs to the gating function in the MoE module share high similarity across successive layers, due to the residual structure of LLMs. As reported in these works, the prediction accuracy can reach about 90\%, significantly enhancing the effectiveness of this approach.
Pre-gated MoE~\cite{hwang2024pre} introduces a pre-gated MoE structure, which identifies the experts required by the next layer during the computation of the current layer and preloads them accordingly. Although this new structure is suitable for prefetching, it may cause a decrease in model accuracy. EdgeMoE~\cite{yi2023edgemoe} constructs a prediction table using a calibrated dataset, leveraging the experts from the current layer to predict the experts for the next layer. This table is built on the observation that the expert activations of sequential layers are statistically correlated. This means that given prior knowledge of the expert activations from layers 1 to n-1, we can estimate the activation probability of each expert at the n-th layer with high confidence.
DyNN-Offload~\cite{dynnoffload} trains a pilot model for each layer to predict the required experts for the next layer. MoE-Infinity~\cite{xue2024moe} tracks the request-level usage frequency of each expert and prefetches high-priority experts based on this frequency. It prefetches experts across multiple layers concurrently, rather than just the next layer, but prioritizes the experts closest to the currently executed layer. ProMoE~\cite{song2024promoe} uses a learned predictor to prefetch experts in a sliding-window manner. This predictor, a small neural network (a two-layer MLP), learns the correlation between layer inputs and expert selections. It can use the inputs of the i-th layer to predict the expert selections of the (i+k)-th layer, where k is the sliding window size (e.g., 4 or 8). ProMoE runs LLM inference for multiple iterations to collect training data for the predictor in the offline stage and runs the predictor on the CPU to hide its execution latency by overlapping it with the LLM inference during the online stage.
Additionally, some works aim to predict all the required experts for the current forward pass at the start of the forward iteration. For example, ExpertFlow~\cite{he2024expertflow} trains a transformer-based predictor, which is particularly well-suited for this task due to its exceptional capabilities in contextual modeling, to predict all the required experts for the current forward pass and prefetch them in advance. This predictor is trained as a classification task, outputting the state of each expert as either active or idle. SiDA~\cite{du2024sida} employs a network-based hash function to predict the activated experts for each token at each layer for the current incoming batch of data. This predictor is constructed with an LSTM and trained using truncated knowledge distillation. Similarly, Read-ME~\cite{readmoe} constructs a pre-gating router decoupled from the MoE backbone to predict all the required experts at once.

\subsubsection{Expert Caching}

Since GPU memory can only hold a subset of experts (the expert cache), optimizing the cache management policy is crucial for improving the cache hit ratio and reducing transfer time. Several works~\cite{eliseev2023fast, yi2023edgemoe, hwang2024pre, yuan2024efficient, huang2023towards} adopt the Least Recently Used (LRU) policy, based on the assumption that recently used experts are more likely to be accessed again. According to Mixtral-8x7B~\cite{jiang2024mixtral}, when an expert is selected for the i-th token, it has a significantly higher probability (more than 10\% in most layers) of being selected again for the (i+1)-th token compared to a random selection.
In contrast, MoE-Infinity~\cite{xue2024moe} employs a variant of the Least Frequently Used (LFU) policy to define the caching priority of experts, leveraging its request-level tracking capability. Fiddler~\cite{kamahori2024fiddler} maintains a set of important experts in GPU memory using a static dataset to profile the active times of experts, treating these times as a measure of their importance. This method demonstrates an improvement of approximately 3 percentage points in hit rate compared to random placement. Similarly, SwapMoE~\cite{kong2023serving} stores a set of important experts in GPU memory based on a defined expert importance score, but it dynamically updates the cached experts at a suitable frequency.
AdapMoE~\cite{zhong2024adapmoe} introduces a dynamic cache size for different model layers, which is driven by its adaptive gating algorithm. This algorithm selects a varying number of experts for different layers, thus adjusting the cache size accordingly. Specifically, it assigns a larger cache size to layers that select more experts or experience lower prediction accuracy. HOBBIT~\cite{tang2024hobbit} proposes a multidimensional cache policy that combines LRU, LFU, and a novel Least High Precision Used (LHU) strategy for its mixed-precision expert cache. Since high-precision experts incur higher loading latencies than low-precision experts, the LHU policy prioritizes high-precision experts to reduce loading costs. Additionally,
CacheMoE~\cite{skliar2024mixture} employs a cache-aware strategy that adjusts the router logits \( z = G(x) \) to enhance the likelihood of selecting experts already present in the cache. This approach effectively increases the expert cache hit ratio, thereby reducing the costs associated with expert loading.

\subsubsection{Expert Loading}

Expert loading time often constitutes the primary bottleneck in expert-offloading inference. As reported in HOBBIT~\cite{tang2024hobbit}, expert loading consumes more than 80\% of the total inference time. To address this, several works aim to directly reduce loading costs. For instance, EdgeMoE~\cite{yi2023edgemoe} and HOBBIT~\cite{tang2024hobbit} reduce loading times by using low-precision experts instead of high-precision ones. Low-precision expert loading requires significantly less time than high-precision experts and can still maintain model accuracy, provided that only the less important experts are replaced.
EdgeMoE determines the precision of each expert by profiling its importance using a static dataset. However, this method is highly dependent on the profiled dataset, which reduces its flexibility across different environments and may negatively affect accuracy. To address this limitation, HOBBIT introduces a token-level dynamic expert loading mechanism. This mechanism dynamically selects the appropriate precision for cache-missing experts based on the current inputs, ensuring both accuracy and flexibility. Specifically, it computes the importance score of a cache-missing expert based on the outputs of the gating module. If the score is below a predefined threshold, the system fetches a low-precision version; otherwise, it retrieves the high-precision version to preserve model accuracy.
Additionally, AdapMoE~\cite{zhong2024adapmoe} employs an adaptive gating algorithm that skips less important experts. This algorithm uses the Fisher information matrix to compute the importance of each expert, further lowering expert loading costs.

\subsubsection{CPU Assisting}
Beyond relying solely on the GPU for computation, some approaches leverage the CPU to assist in the process. Fiddler~\cite{kamahori2024fiddler} performs MoE computation on the CPU by copying intermediate activation values from GPU memory to CPU memory when the required experts are unavailable in GPU memory and then transferring the outputs back to the GPU. This method eliminates the need to load missing experts into GPU memory. As measured, the latency for loading an expert from CPU memory to GPU memory is significantly higher than the combined latency of copying activation values to the CPU, performing expert computation on the CPU, and then copying the results back to the GPU.
HOBBIT~\cite{tang2024hobbit} follows a similar pattern, using the CPU for computation with low-precision experts. Additionally, MoE-Lightning~\cite{cao2024moelight} introduces an innovative CPU-GPU-I/O pipelining strategy that enables the simultaneous utilization of both CPU and GPU resources, thus enhancing overall system efficiency. However, this approach is only applicable when CPU resources are sufficiently available. It may not be suitable for platforms with limited CPU resources or shared memory devices, such as the Jetson Orin.

\begin{table}[]
\resizebox{\textwidth}{!}{
\begin{tabular}{ccccl}
\toprule
Method             & Speedup                    & Metric & Baseline              & Main Techniques                             \\ \midrule
ProMoE~\cite{song2024promoe}             & 1.16x                      & Tokens per second           & LRU-MoE~\cite{song2024promoe}            & Train a predictor to prefetch experts       \\ 
Fiddler~\cite{kamahori2024fiddler}            & 8.20x                       & Tokens per second           & Mixtral-Offloading~\cite{eliseev2023fast}   & Use CPU to help computation                 \\ 
HOBBIT~\cite{tang2024hobbit}             & 2.30x                       & Tokens per second           & MoE-Infinity~\cite{xue2024moe}            & Load adaptive precision experts           \\  
AdapMoE~\cite{zhong2024adapmoe}            & 1.36x & Latency per sample          & Mixtral-Offloading~\cite{eliseev2023fast}    &  Skip  unimportant experts  dynamically     \\ 
ExpertFlow~\cite{he2024expertflow}         & 1.99x                      & Tokens per second           & Cache-MoE~\cite{he2024expertflow}        & Train a predictor to prefetch experts       \\
MoE-Infinity~\cite{xue2024moe}       & 1.96x                      & Latency per token           & Mixtral-Offloading~\cite{eliseev2023fast}       & Fetch experts with request-level tracing \\ 
MoE-Lightning~\cite{cao2024moelight}      & 3.50x                       & Tokens per second           & FlexGen~\cite{sheng2023flexgen}              & Schedule CPU-GPU-I/O pipeline              \\ 
Mixtral-Offloading~\cite{eliseev2023fast} & 2.28x & Tokens per second           & Accelerate~\cite{transformers}   & Use LRU to cache important experts          \\ \bottomrule
\end{tabular}
}
\caption{Speedup of offloading systems evaluated on Mixtral-8x7B model~\cite{jiang2024mixtral}.}
\label{tab:mixtral-offloading}
\end{table}

\begin{table}[]
\resizebox{\textwidth}{!}{
\begin{tabular}{ccccl}
\toprule
Method        & Speedup & Metric & Baseline          & Main Techniques                                         \\ \midrule
EdgeMoE~\cite{yi2023edgemoe}       & 2.82x   & Latency per token           & IO-EXP~\cite{yi2023edgemoe}        & Quantize experts into different precisions              \\ 
SwapMoE~\cite{kong2023serving}       & 2.00x      & Latency per sample          & Original-MoE~\cite{kong2023serving}     &  Update cached experts dynamically  \\ 
SIDA~\cite{du2024sida}      & 3.93x   & Samples per second          & DeepSpeed~\cite{deepspeed}     & Prefetch experts with a hash function     \\ 
ExpertFlow~\cite{he2024expertflow}    & 5.23x   & Tokens per second           & SE-MoE~\cite{shen2022se}    & Train a predictor to prefetch experts                   \\ 
MoE-Infinity~\cite{xue2024moe}  & 4.53x   & Latency per token           & Accelerate~\cite{transformers}  & Prefetch experts with request-level tracing             \\ 
Pre-gated MoE~\cite{hwang2024pre} & 1.50x    & Tokens per second           & MoE-OnDemand~\cite{hwang2024pre}  & Design a pre-gate MoE structure               \\ \bottomrule
\end{tabular}
}
\caption{Speedup of offloading systems evaluated  on Switch Transformer model~\cite{fedus2022switch}.}
\label{tab:stmoe-offloading}
\end{table}

\subsubsection{Performance Summarization}

Mixtral-8x7B~\cite{jiang2024mixtral} and the Switch Transformer~\cite{fedus2022switch} are popular MoE models, and most offloading systems use them as evaluation workloads. We extract relevant performance data from various papers to assess their effectiveness. Table~\ref{tab:mixtral-offloading} shows that many studies use Mixtral-Offloading as the baseline when evaluating the Mixtral-8x7B model. From the table, it is evident that Fiddler achieves a significant speedup compared to other systems, as it effectively utilizes the CPU to assist computation in GPU-limited environments. However, Table~\ref{tab:stmoe-offloading} shows that the methods when evaluated on Switch Transformer use different baselines, making direct comparisons difficult. Therefore, for our own studies on this topic, Mixtral-8x7B should stand out as a more suitable workload to consider.

\subsection{Framework}

\begin{table}[]
\centering
\resizebox{\textwidth}{!}{
\begin{tabular}{cccccccc}
\toprule
                          & Pytorch~\cite{pytorch} & Transformers~\cite{transformers} & DeepSpeed~\cite{deepspeed} & Fairseq~\cite{fairseq} & Llama.cpp~\cite{llamacpp} & vLLM~\cite{kwon2023efficient} & FasterTransformer~\cite{FasterTransformer} \\ \midrule
Parallelism System & 12      & 5            & 9         & 2       & 0         & 0    & 0                        \\ 
Offloading System  & 4       & 7            & 1         & 0       & 2         & 1    & 1                        \\ \bottomrule
\end{tabular}
}
\caption{Statistics of the number of systems built on popular current frameworks.}
\label{tab:framework}
\end{table}

Most of the systems mentioned above are built on top of popular deep learning frameworks such as PyTorch~\cite{pytorch}, Transformers~\cite{transformers}, and DeepSpeed~\cite{deepspeed}. The data is summarized in Table~\ref{tab:framework}. From the table, it is evident that most works in the expert parallelism category implement their systems based on PyTorch, DeepSpeed, and Transformers. PyTorch serves as the foundational deep learning framework, with other libraries like Transformers and DeepSpeed built on top of it. As a result, directly building a system based on PyTorch offers greater flexibility in designing custom functions, though it lacks some popular techniques found in other libraries, such as model parallelism. DeepSpeed provides numerous features to assist in model training, particularly in distributed environments. Transformers, on the other hand, offers many pretrained models, allowing developers to easily modify the model structure within the library. While this is very convenient for quickly testing ideas, it may not provide the best performance in terms of throughput and latency.

In the expert offloading category, most works still rely on Transformers due to its ease of use. However, some works also utilize Llama.cpp~\cite{llamacpp}, vLLM~\cite{kwon2023efficient}, and FasterTransformers~\cite{FasterTransformer}. These three frameworks are optimized for deploying LLMs and offer great performance, but they tend to be more complex to use.

In conclusion, if you are looking to quickly verify an idea, Transformers is an excellent choice. For training models in distributed environments, DeepSpeed is the better option. If your focus is on designing an inference system with high performance, Llama.cpp and vLLM are good starting points. Otherwise, PyTorch remains a solid choice.

\section{Hardware-Level Optimization}\label{sec:hardware}

Recent hardware optimizations for MoE inference have addressed key challenges through novel architectures and co-design approaches. These optimizations target critical issues such as operations per byte (Op/B) efficiency, heterogeneous computing units, and memory access patterns. The following discusses significant advances in hardware-level solutions.

\begin{figure}[t]
    \centering
    \subfloat[Overview of MoNDE]{\includegraphics[width=0.618\linewidth]{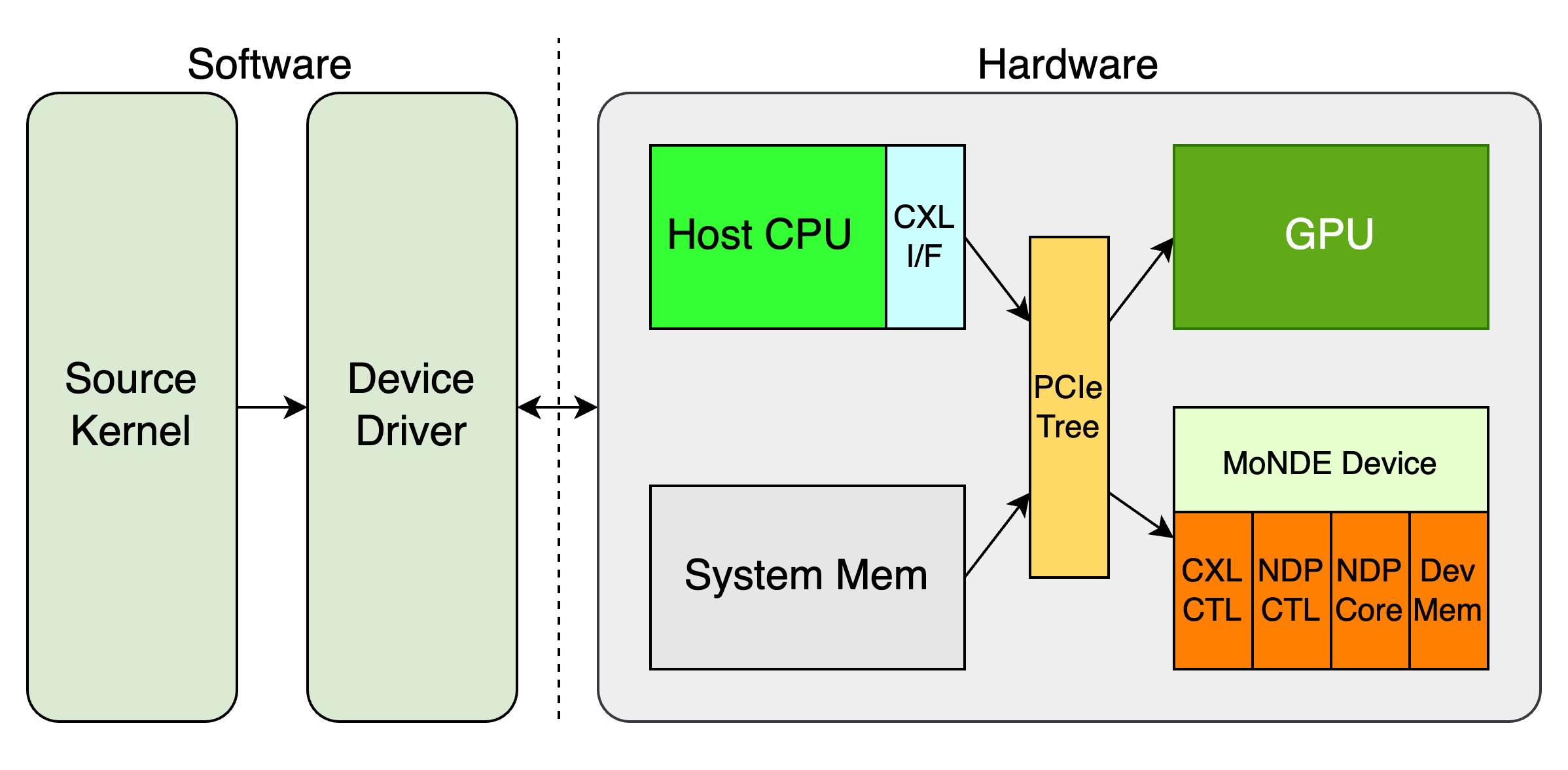}}
    \subfloat[Parameter and Activation Movement]{\includegraphics[width=0.382\linewidth]{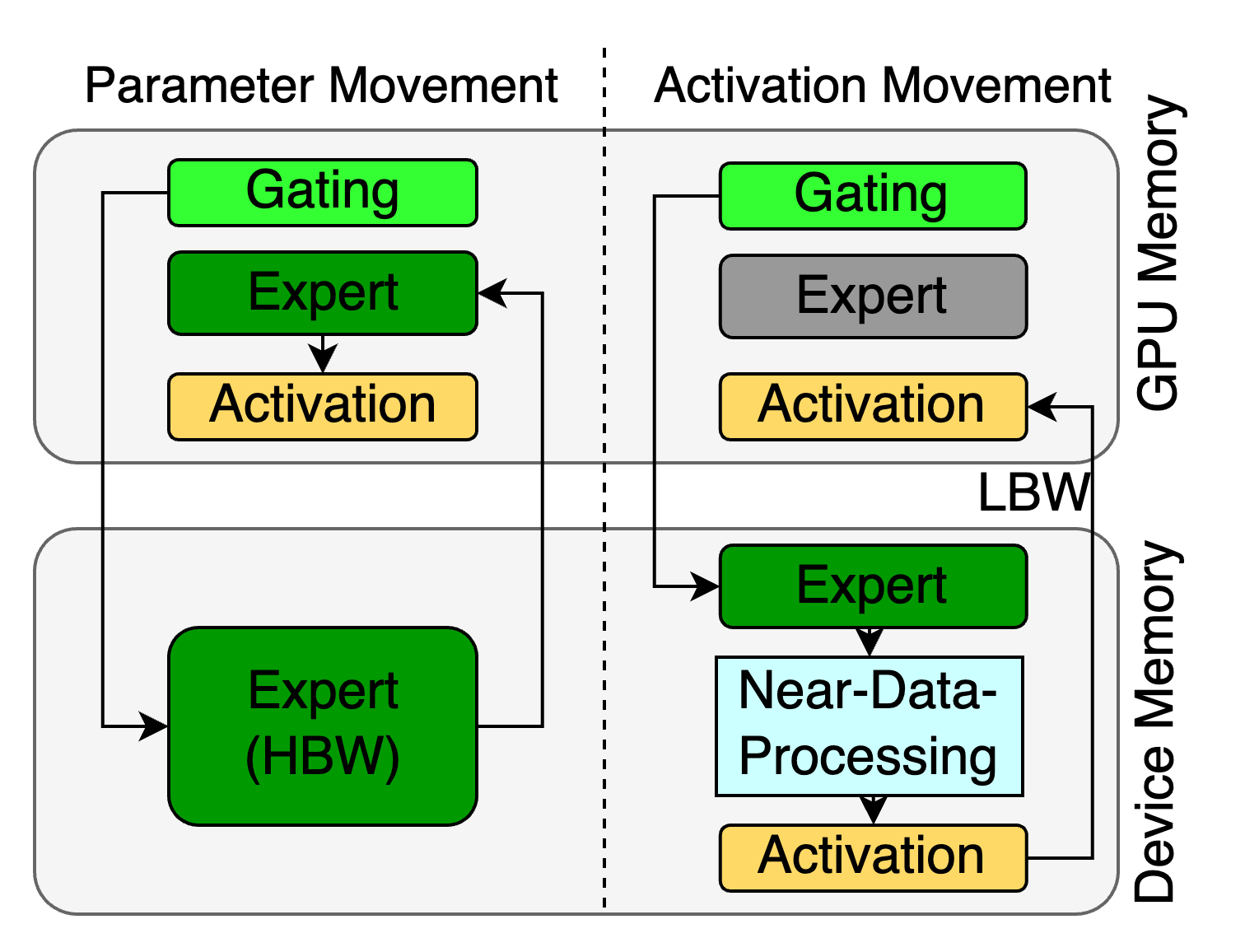}}
    \caption{(a) MoNDE~\cite{Kim2024monde} system design integrating software and hardware. (b) the data flow between AM and PM.}
  \label{fig:monde}
\end{figure}

MoNDE~\cite{Kim2024monde} introduces a near-data processing (NDP) solution to address the challenges of sparse activation and expert parameter transmission overhead (Fig. \ref{fig:monde}). The architecture integrates CXL (Compute Express Link)-based NDP controllers with specialized NDP cores for in-memory computation, utilizing LPDDR SDRAM  (low power double data rate synchronous dynamic random-access memory) for high bandwidth and energy efficiency. The system implements a hybrid computing strategy where GPUs handle frequently accessed "hot" experts while NDP units process "cold" experts, enabling concurrent execution through an Activation Movement paradigm rather than traditional Parameter Movement.

FLAME~\cite{Lin2024flame} is the first accelerating framework that fully exploits MoE sparsity for transformers on FPGA. At the parameter level of the model, it utilizes M:N pruning to reduce unnecessary calculations, which can make a balance between column-balanced structured pruning and unstructured pruning. At the expert level, sparse activation prediction is performed through CEPR(circular expert prediction). By changing the patterning of the activation path of experts, the accuracy of expert predictions can be effectively improved. Then use a double buffering mechanism to load the predicted expert while computing the previous expert to improve expert deployment efficiency.

M\textsuperscript{3}ViT~\cite{Liang2022m3vit} and Edge-MoE~\cite{Sarkar2024edgemoe} constructed their FPGA architecture based on the reordering of attention computation in multitasking scenarios. For inference, M\textsuperscript{3}ViT can activate only the sparse "expert" pathway relevant to the task of interest for efficiency, and can further achieve zero-overhead switching between tasks with their hardware-level co-design. Edge-MoE is the first end-to-end FPGA implementation for multi-task ViT proposed some aggressive techniques, including an approximate method for solving the excessive complexity of GELU function calculation on FPGA and a unified linear layer module to achieve efficient reuse of hardware resource.

Duplex~\cite{Yun2024Duplex} selects a destination suitable for each layer execution in the device that combines xPU and Logic PIM (processing-in-memory). That means it could integrate two types of processing units that share device memories. Due to the bottleneck in computation and memory access between these two processing units which can complement each other, high computation and memory access utilization can be achieved simultaneously on the same device. Besides, it introduced an alternative PIM microarchitecture. The logic PIM optimized low Op/B operations through powerful processing units on the logic die, as well as more through silicon vias (TSVs) to achieve high bandwidth communication between the DRAM die and the logic die. Furthermore, it could execute expert and attention stages in parallel to maximize inference efficiency.

Space-mate~\cite{Park2024spacemate} has provided its accelerator design for SLAM (simultaneous localization and mapping) tasks on mobile devices. Mainly including an Out-of-Order (OoO) SMoE router to alleviate data transactions for low latency, a single skip (SS) and dual-skip (DS) heterogeneous core architecture to exploit coarse-grained sparsity caused by similar zero patterns in the same expert for high throughput and energy-efficiency.

\section{Future Directions and Open Challenges}\label{sec:future}

Despite significant advances in MoE inference optimization, critical challenges and opportunities exist throughout the computing stack. This section presents a systematic analysis of future research directions, organized along three fundamental dimensions: (1) computing infrastructure, from hardware to system software, (2) key system requirements, including performance, reliability, and efficiency, and (3) development support ecosystem. These dimensions are interconnected: advances in infrastructure enable better system properties, while development tools accelerate progress in both areas.

\subsection{Computing Infrastructure Optimization}

\subsubsection{Hardware Integration and Acceleration}

The efficient execution of MoE models demands novel hardware architectures and acceleration strategies that go beyond traditional computing paradigms~\cite{Sarkar2024edgemoe, Liang2022m3vit, Yun2024Duplex, Park2024spacemate}. Current hardware platforms, optimized primarily for dense computations, often struggle to efficiently handle the sparse and dynamic nature of MoE workloads~\cite{sun2022vaqf, you2023vitcod, 9773212}. This necessitates the development of specialized hardware solutions that can better support the unique computational patterns of MoE inference.

Traditional hardware optimization presents several immediate opportunities for improvement. The development of specialized circuits for expert routing and activation could significantly reduce the overhead of dynamic expert selection. Memory hierarchies optimized for sparse parameter access could better support the irregular access patterns characteristic of MoE computation. Furthermore, the implementation of efficient hardware support for dynamic workload patterns could enhance the overall system performance.

Emerging hardware platforms offer exciting new possibilities for MoE acceleration. Neuromorphic computing systems~\cite{schuman2017survey, schuman2022opportunities}, with their inherent support for sparse, event-driven computation, could provide natural acceleration for expert activation patterns. Quantum computing platforms~\cite{liang2023unleashing, steane1998quantum, padmanaban2024quantum} might enable novel approaches to complex routing decisions, though significant challenges remain in effectively bridging classical and quantum computations. The integration of novel memory technologies could also provide more efficient storage and access patterns for expert parameters.

\subsubsection{System Software Optimization}
The optimization of the system software stack presents significant challenges for MoE models due to their unique computational patterns and resource requirements. Current operating systems and runtime environments, designed primarily for traditional dense neural networks, lack native support for efficient sparse computation and dynamic expert scheduling~\cite{Brainstorm}. This gap between hardware capabilities and application requirements necessitates fundamental innovations in system software design.

Memory management systems need a significant redesign to handle the dynamic nature of expert activation. Traditional virtual memory systems and cache hierarchies for LLM are optimized for regular access patterns, while MoE models exhibit highly irregular, input-dependent memory access behaviors~\cite{kwon2023efficient}. Future research must develop specialized memory management techniques that can predict and prefetch expert parameters more effectively, potentially incorporating new abstractions for sparse tensor operations.

Resource allocation in distributed environments requires particular attention. The dynamic nature of expert activation makes it difficult to optimize placement and migration decisions~\cite{10319949, cong2024prediction}. Future systems need to incorporate robust mechanisms for load balancing, fault tolerance, and resource elasticity. This includes developing intelligent middleware that can adapt to changing workload patterns and system conditions while maintaining efficient resource utilization.

The co-design of hardware and software components represents a crucial challenge in this domain, reflecting the fundamental gap between traditional computing architectures designed for dense computation and the unique requirements of MoE models. Future research must focus on developing hardware architectures that can minimize energy consumption while maintaining performance, including exploring novel approaches to reducing data movement and developing hardware-level techniques for trading off accuracy for efficiency. Success in this area requires close collaboration between hardware architects and algorithm designers to ensure that hardware optimizations align with the computational requirements of MoE models, as progress in both hardware and software layers, and particularly their co-design, is essential for efficient MoE deployment.

\subsection{Operational Challenges and Requirements}
The deployment and operation of MoE systems in production environments introduce complex challenges that span performance optimization, system reliability, and resource efficiency. These operational requirements often interact and sometimes conflict, requiring careful balance in real-world settings. Understanding and addressing these challenges is crucial for the practical adoption of MoE systems across different application domains and deployment scenarios.

\subsubsection{Energy Efficiency and Sustainability}
Energy efficiency and environmental impact considerations have received relatively less attention in MoE inference optimization research, which has predominantly focused on throughput and latency metrics. As AI systems continue to scale and their environmental impact becomes more significant, there is a pressing need to consider energy consumption and carbon emissions as primary optimization objectives alongside traditional performance metrics~\cite{mao2024green, cruz2024innovating, raman2024green, alizadeh2024green, hou2024improving}.

The energy consumption patterns of MoE models present unique challenges due to their sparse activation patterns and distributed nature. While sparse activation theoretically reduces computational demands, the overhead from expert routing, communication, and load balancing can lead to significant energy costs~\cite{hwang2023tutel, nie2023flexmoe, 10.1145/3627703.3650083}. Current hardware platforms, often optimized for dense computations, result in suboptimal energy efficiency when handling the dynamic workloads characteristic of MoE inference~\cite{zhu2024lightening, ning2024photonic}. Additionally, the distributed nature of many MoE deployments introduces substantial energy overhead from data movement and communication.

Carbon-aware deployment strategies represent an important direction for future research~\cite{faiz2023llmcarbon, radovanovic2022carbon, fu2024llmco2}. In distributed settings, expert placement and workload distribution decisions should consider not just computational resources but also the carbon intensity of different data centers' power sources. This could involve developing dynamic scheduling algorithms that preferentially route computation to locations with access to renewable energy. Understanding and optimizing the trade-off between energy consumption and model quality is crucial, requiring methods to quantify these relationships and creating adaptive mechanisms that can dynamically adjust this trade-off based on application requirements and energy constraints.

The development of comprehensive energy accounting frameworks is essential for accurate optimization. Current evaluation methodologies often fail to account for the full energy cost of MoE inference, including both direct computational costs and indirect costs from data movement, cooling, and infrastructure overhead. Future research must focus on developing more sophisticated energy models that can capture these various aspects of energy consumption, enabling more informed optimization decisions that consider the full environmental impact of MoE deployment.

\subsubsection{Latency and Quality of Service}
The dynamic and distributed nature of MoE inference introduces significant challenges in maintaining consistent performance and reliability. Unlike traditional neural networks with fixed computation patterns, MoE models exhibit variable execution paths and resource requirements, making it difficult to provide consistent latency guarantees and maintain high availability~\cite{Brainstorm, xue2024moe}.

Predictable expert activation and routing represent a fundamental challenge. The input-dependent nature of expert selection can lead to significant variability in processing time and resource utilization. Research is needed to develop techniques that can better predict and manage this variability, potentially through advanced caching strategies, workload characterization, and adaptive routing algorithms~\cite{zhong2024adapmoe, song2024promoe, he2024expertflow}. This includes methods for balancing the trade-off between routing accuracy and computational overhead.

System reliability and fault tolerance require particular attention in distributed MoE deployments~\cite{cai2024mocsystemefficientfaulttolerance}. The failure of individual experts or communication links can significantly impact system performance and model quality. Research opportunities include developing robust failure detection and recovery mechanisms, implementing graceful degradation strategies, and designing redundancy schemes that balance reliability with resource efficiency. This includes methods for maintaining service quality even when some experts are unavailable or performing sub-optimally.

\subsection{Development Support Ecosystem}

\subsubsection{Open-Source Frameworks}
Current deep learning frameworks like PyTorch~\cite{pytorch} and TensorFlow~\cite{TensorFlow} lack native optimizations for MoE workloads, treating them as an auxiliary use case rather than a primary optimization target. This gap between framework capabilities and MoE requirements creates significant challenges for researchers and practitioners attempting to develop and deploy MoE systems efficiently.

Core framework support for MoE-specific operations requires fundamental enhancements. Research opportunities include developing specialized operators for expert routing, implementing efficient sparse computation primitives, and creating optimized memory management systems for expert parameters~\cite{qian2024eps, GatingDropout}. These enhancements must be deeply integrated into framework internals to achieve performance comparable to dense model optimization.

High-level APIs and abstractions are essential for making MoE development accessible to a broader community. Current frameworks often require significant expertise to implement MoE architectures efficiently, limiting adoption and experimentation~\cite{llamacpp}. Research needs include developing intuitive APIs for expert definition and routing configuration, implementing automated optimization passes for MoE workloads, and creating deployment abstractions that handle distributed execution complexity. These APIs must balance ease of use with the flexibility to implement novel MoE architectures and optimization strategies.

Framework integration with existing ML ecosystems represents another critical challenge. MoE systems must coexist with traditional models and leverage existing tools for training, debugging, and deployment. Future research should focus on developing standardized interfaces for MoE components, implementing compatibility layers with popular ML tools, and creating unified optimization pipelines that can handle both dense and sparse computation patterns. This integration must preserve framework performance while maintaining compatibility with the broader ML ecosystem.

\subsubsection{Benchmarking and Standardization}

The rapid advancement of MoE inference optimization techniques has created a pressing need for comprehensive benchmarking and standardization frameworks~\cite{nguyen2024libmoe, fu2024moe}. Although many LLM benchmarks have been developed in recent years~\cite{Llm-perf, Llm-Artificial, reddi2020mlperf,xu2023mmbench}, they are not well-suited for evaluating MoE systems due to a lack of consideration for the unique characteristics of the MoE architecture~\cite{fu2024moe}. Current evaluation practices often vary significantly across studies, making it difficult to perform fair comparisons and assess the relative merits of different optimization approaches. This fragmentation in evaluation methodologies impedes the systematic progress of the field and complicates the adoption of optimization techniques in practical applications.

A standardized benchmarking framework should encompass multiple dimensions of MoE inference optimization. This includes performance metrics such as latency, throughput, and resource utilization, as well as model quality metrics across different tasks and domains. The framework should also consider various deployment scenarios, from edge devices to large-scale distributed systems, and different hardware configurations. Furthermore, standardized workload patterns that reflect real-world usage scenarios are essential for meaningful evaluation of optimization techniques.

The development of such benchmarking standards presents several challenges. The dynamic nature of MoE models, where different inputs may activate different experts, makes it difficult to establish consistent performance measurements. The trade-offs between various optimization objectives (e.g., latency vs. throughput, or quality vs. efficiency) need to be carefully balanced in the evaluation criteria. Moreover, the diverse hardware platforms and deployment scenarios used in practice require benchmarks that can meaningfully capture performance across different contexts.

Looking forward, the field would benefit from the establishment of standardized benchmark suites and consistent evaluation methodologies. The development of reference implementations and baseline systems would facilitate comparative analysis, while the definition of representative workload patterns would ensure relevance to practical applications. These standardization efforts would ultimately accelerate the advancement of MoE inference optimization research by enabling more rigorous and comparable evaluations of new techniques.

\section{Conclusion}\label{sec:conclusion}
This survey provides a comprehensive overview of inference optimization techniques for MoE models across different abstraction levels. We have systematically analyzed various approaches, from model-level optimizations including efficient expert design, compression techniques, and algorithm improvements, to system-level solutions for distributed computing and expert offloading, and finally to hardware-level acceleration designs. Through this analysis, we observe that while MoE models offer a promising approach to scale model capacity with controlled computational costs, their efficient deployment requires careful consideration and optimization at multiple levels of the system stack. The field has seen rapid advancement with numerous innovative solutions, yet several challenges remain, particularly in areas such as hardware integration, energy efficiency, and standardized benchmarking.

Looking forward, we anticipate continued evolution in MoE inference optimization, driven by both academic research and industrial applications. The increasing adoption of MoE architectures in large-scale language models and multimodal systems will likely spur further innovations in optimization techniques. Key areas for future research include the development of specialized hardware architectures for sparse computation, more efficient expert routing algorithms, and improved system-level solutions for distributed deployment. We hope this survey serves as a valuable reference for researchers and practitioners working on MoE deployment, providing a structured understanding of current approaches and inspiring new directions in this rapidly evolving field.

\bibliographystyle{acm}
\bibliography{sample-base}
\end{document}